\pdfoutput=1
\documentclass[11pt]{article}
\newif\ifreview
\reviewfalse

\usepackage[final]{acl}
\usepackage{comment}
\usepackage{times}
\usepackage{latexsym}
\usepackage{multirow}
\usepackage{xcolor}
\usepackage[T4,T1]{fontenc}
\usepackage{textgreek}
\usepackage[utf8]{inputenc}
\usepackage{tipa}
\usepackage{microtype}

\usepackage{inconsolata}

\usepackage[utf8]{inputenc}
\usepackage[T1]{fontenc}
\usepackage{tipa}          % IPA symbols
\usepackage{newunicodechar}
\usepackage{graphicx}
\usepackage[T1]{fontenc}
\usepackage{enumitem}
\usepackage{tcolorbox}
\tcbuselibrary{skins}
\usepackage{amsmath}
\usepackage{float}
\usepackage[dvipsnames]{xcolor}
\usepackage{calc}
\usepackage{graphicx}

\usepackage{pdflscape}
\usepackage{longtable} 
\usepackage{booktabs}
\usepackage{siunitx}
\usepackage{adjustbox}

\usepackage{tikz}
\usetikzlibrary{arrows.meta, positioning}

\usepackage{caption}
\usepackage[table]{xcolor}
\usepackage{soulutf8}

\usepackage{todonotes}

\definecolor{tticblue}{RGB}{0, 94, 184}

\definecolor{Scolor}{RGB}{255,230,230}   % salmon
\definecolor{Gcolor}{RGB}{230,230,255}   % violet
\definecolor{KPcolor}{RGB}{230,255,230}  % green
\definecolor{Tcolor}{RGB}{255,255,230}   % yellow

\newcommand{\SHL}[1]{{\sethlcolor{Scolor}\hl{#1}}}
\newcommand{\GHL}[1]{{\sethlcolor{Gcolor}\hl{#1}}}
\newcommand{\KPHL}[1]{{\sethlcolor{KPcolor}\hl{#1}}}
\newcommand{\THL}[1]{{\sethlcolor{Tcolor}\hl{#1}}}

\title{\textsc{LingGym}: How Far Are LLMs from Thinking Like Field Linguists?}

\author{Changbing Yang\textsuperscript{\textipa{\OE}}, Franklin Ma\textsuperscript{\textipa{\OE}}, Freda Shi\textsuperscript{\textipa{W},\textipa{V}}, Jian Zhu\textsuperscript{\textipa{\OE}} \\
\textsuperscript{\textipa{\OE}}University of British Columbia\\
\textsuperscript{\textipa{W}}University of Waterloo \quad 
\textsuperscript{\textipa{V}}Vector Institute \\
  \texttt{cyang33@mail.ubc.ca}, \texttt{franklin.ma@ubc.ca}\\
  \texttt{fhs@uwaterloo.ca}, \texttt{jian.zhu@ubc.ca}
  }

\begin{document}
\maketitle
\begin{abstract}
This paper introduces \textsc{LingGym}, a new benchmark that evaluates LLMs’ capacity for meta-linguistic reasoning using Interlinear Glossed Text (IGT) and grammatical descriptions extracted from 18 typologically diverse reference grammars. Unlike previous work that focuses on specific downstream tasks, we assess whether LLMs can generalize linguistic inference across low-resource languages and structures not seen during training. We present a controlled evaluation task: Word-Gloss Inference, in which the model must infer a missing word and gloss from context using varying levels of linguistic information (e.g., glosses, grammatical explanations, translations). Our results show that incorporating structured linguistic cues leads to consistent improvements in reasoning performance across all models. This work highlights both the promise and current limitations of using LLMs for typologically informed linguistic analysis and low-resource language documentation.
%Large language models (LLMs) have demonstrated impressive performance on several multilingual tasks; however, they often suffer when translating texts to and from low-resource languages because there is little to no data for them in comparison to the abundance of data used to train high-resource languages (e.g. English, Mandarin). With thousands of low-resource languages facing extinction, it has become increasingly crucial to examine the extent of linguistic bias present in these LLMs and to investigate ways we can counter such effects. We present a high-quality, cleaned dataset of Interlinear Glossed Text (IGT) from reference grammars available in the \textit{Language Science Press} journal. We experiment with a single language and demonstrate improved \texttt{chrF} scores from providing grammatical information into the prompt compared to an open-source baseline. We hope this dataset can enable further research into enhancing multilingual support in LLMs for a wider range of underrepresented languages.
\end{abstract}

\section{Introduction}
In recent years, researchers have been actively exploring how large language models \citep[LLMs;][]{openai2024gpt4technicalreport, qwen2.5, grattafiori2024LLaMA3herdmodels} can assist and accelerate scientific discoveries in various disciplines \cite[e.g.,][]{romera2024mathematical,merchant2023scaling,fawzi2022discovering,hayes2025simulating,zhang2024scientific}. 
However, exploration on how LLMs can assist social sciences is relatively limited \cite{grossmann2023ai,bail2024can,ziems2024can}. 
In particular, LLMs with the capacity of reasoning about meta-linguistic knowledge have the potential to become powerful tools for language documentation, linguistic hypothesis testing, and typological research. 
For example, by generalizing linguistic structures such as morphology, syntax, and word order across languages, they can suggest morpheme segmentations and glosses, identify patterns or counterexamples to test hypotheses, and compare structural features across different languages.

On the other hand, while recent advances have shown impressive performance in high-resource languages like English, our understanding of their effectiveness on typologically diverse and underrepresented languages remains limited \cite{alhanai2025bridging}, mainly due to the overwhelming dominance of English and other high-resource languages in their training data \cite{blasi-etal-2022-systematic, khade-etal-2025-challenges, li2024languagerankermetricquantifying, wu-dredze-2020-languages}. 

\begin{figure}[t]
  \centering
  \begin{tcolorbox}[
    enhanced,
    colback=white,
    colframe=black,
    arc=4mm,
    boxrule=0.5pt,
    width=\columnwidth
  ]
  \hspace{-10pt}
  \begin{tabular}{p{0.98\columnwidth}}
    \vspace{-10pt}
    \textbf{Predicative adjectives} \\[2pt]
    When adjectives function predicatively, they may receive copular morphology, although this is not obligatory (neither for adjectives nor for nouns).
These predicative adjectives occur clause-finally (the position held prototypically
by verbs). 
    \vspace{5pt} \\
    \hline \\
    \vspace{-18pt}
    \textbf{Orthography}: \textit{Mï \textcolor{red}{anmapïna}.} \\[-6pt]
    \textbf{Segmentation:} mï anma=p-na\\[2pt]
    \textbf{Gloss:} \textsc{3sg.subj} good=\textsc{cop-irr} \\[2pt]
    \textbf{Translation:} `It will be good.' \\[-2pt]
  \end{tabular}
  \end{tcolorbox}
  \vspace{-10pt}
  \caption{An excerpt explaining predicative adjectives in Ulwa, with an associated example as IGT from \textit{A Grammar of Ulwa} \citep[\textit{Papua New Guinea};][p.~166]{ulwa}. The example IGT is represented with the Leipzig Glossing rules \cite{leipzig_glossing_rules}. The text in red highlights the emphasized word discussed in the grammar explanation. The gloss consists of a third-person singular subject marker (\textsc{3sg.subj}) for ``it,''  and an irrealis copula (\textsc{cop-irr}) marker for ``will be.''
  \vspace{-10pt}
  }
  \label{fig:predicativeAdjectives}
\end{figure}

To explore how well LLMs can understand low-resource languages when provided with structured linguistic input, we turn to reference grammars \cite{mosel2006grammaticography,chelliah2013fieldwork}, which aim to comprehensively describe the structure of individual languages. 
Reference grammars offer two valuable types of information:

\begin{enumerate}[leftmargin=*,itemsep=0pt,topsep=4pt]
\item \textbf{Interlinear glossed text (IGT)}, a standard text format used by field linguists to present linguistic data, which is useful for tasks like morphological analysis, syntactic structure identification. 
IGT typically consists of four lines: a phonological or orthographic transcription, a segmentation of words into morphemes, corresponding grammatical glosses, and a free English translation. 
Conventions include hyphens to mark morpheme boundaries, equals signs for clitic boundaries, and periods to separate multiple glossing elements for a single morpheme \cite{leipzig_glossing_rules}, as illustrated in Figure~\ref{fig:predicativeAdjectives}. 
\item \textbf{Grammatical terms and explanations} embedded throughout the text, where important linguistic terms (e.g., tense markers, case particles, verb classes) are defined and contextualized within the grammar. 
\end{enumerate}
Together, these resources reflect the approach taken by human linguists, who analyze unfamiliar languages by studying structured descriptions rather than relying on raw corpora.
Thanks to decades of documentation efforts, such materials are available for many endangered and low-resource languages, presenting a valuable opportunity to test LLMs’ ability to reason over structured linguistic knowledge curated by experts.
Unlike the unstructured web-scale corpora typically used to train LLMs, descriptive grammars hold a unique advantage by offering systematic and interpretable accounts of a language’s morphology and syntax. 
In addition to serving human language learners and linguists, these structured frameworks that encode rich meta-linguistic knowledge also offer a valuable resource for evaluating LLMs. 
By drawing on this explicit information, we can design targeted evaluation tasks that probe model performance across diverse linguistic phenomena and typological patterns.

\begin{figure*}[t]
\centering
  \includegraphics[width=\textwidth]{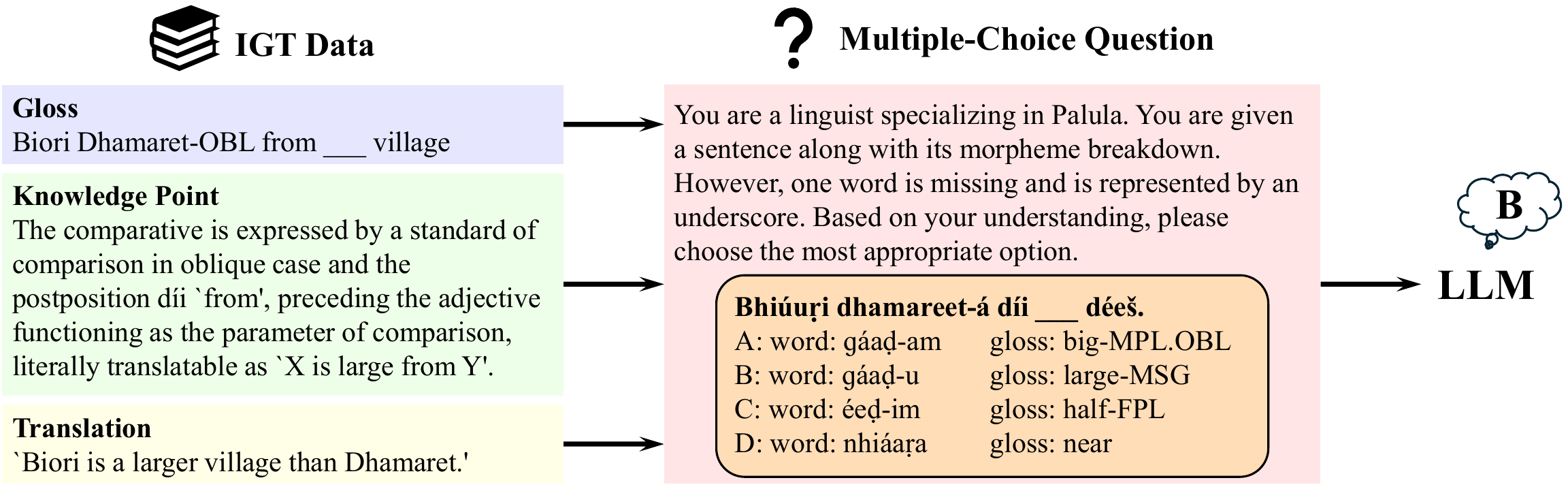}
  \caption{
  An illustration of how IGT data is transformed into a multiple-choice question for evaluating an LLM.
  One word is masked with an underscore in the provided sentence. 
  An LLM takes the constructed prompt and the most contextually appropriate answer based on the linguistic information.
  }
  \label{fig:pipeline}
\end{figure*}

%A key challenge, however, lies in integrating these resources into LLMs. Input length constraints make it impossible \textcolor{red}{it's not impossible for all models}to feed the entire grammar book content into a model. To address this, 
In this work, we design a task-oriented approach: for each target sentence, the LLM receives the utterance or the utterance paired with its glosses, augmented by targeted grammatical cues (e.g., rules about verb conjugation or case marking).
We evaluate the model’s comprehension through a controlled task (Figure \ref{fig:pipeline}): \textbf{word-gloss inference}, where the model anwsers a multiple-choice question to infer a missing word or its corresponding gloss based on the linguistic context.

Our contributions are as follows: first, we present a cleaned and structured dataset of IGT examples drawn from 18 endangered and low-resource languages---these examples are extracted from publicly available reference grammars and subsequently verified by hand (\S\ref{subsec:preprocessing}, \S\ref{sec:lingGymDataset}).
Second, we develop an evaluation framework grounded in descriptive linguistic resources to assess how well LLMs can interpret and infer in low-resource languages using IGT data and grammatical rules (\S\ref{subsec:questionGeneration}). Third, we benchmark multiple state-of-the-art LLMs on our proposed tasks and provide a typologically informed analysis of their performance, highlighting both capabilities and limitations when processing structured linguistic knowledge (\S\ref{subsec:experiments}, \S\ref{subsec:results}). 
We release the benchmark on GitHub.\footnote{
\ifreview
\texttt{Anonymous URL}
\else
\url{https://github.com/changbingY/LingGym}
\fi
}

\section{Related Work}
\subsection{NLP for Low-Resource languages}
The value of language models as tools to assist language documentation and revitalization has been well recognized \cite{bird2020decolonising} in both linguistics and natural language processing (NLP) communities. 
These models enable a variety of applications, including automatic transcriptions of speech \cite{dunbar2017zero,li2020universal,samir-etal-2025-comparative,zhu-etal-2025-zipa}, low-resource speech synthesis \cite{kazantsevaa2024speech,wang-etal-2025-developing}, automatic interlinear glossing \cite{moeller2020igt2p,he-etal-2024-wav2gloss,yang-etal-2024-multiple}, grapheme-to-phoneme conversion \cite{li2022zero,zhu22_interspeech}, and more \cite{gessler2024nlp}. 
Most existing work formulates a specific subtask in language documentation as an established NLP task with standard evaluation metrics. 

While these directions have led to many low-resource NLP technologies, there are still many limitations \cite{bird2020decolonising}. 
First, many models are trained on specific languages where training data is available, and are usually not generalizable to unseen languages. 
Second, many technologies are developed in highly artificial settings with well-defined tasks and clean data. 
As a result, they are unable to solve many linguistic tasks in real language documentation that are more complex, noisy, and subjective.\footnote{Language documentation is often subjective because linguists have different habits, preferences, and theoretical approaches to analysis, and there is no universally standardized method for representing or annotating linguistic data.} 
To bridge the gap, in this work, we present a benchmark in real language documentation scenarios and use it to assess the reasoning capabilities of general-purpose LLMs.
%\textcolor{orange}{[Freda: Please check the appropriateness of this added sentence and feel free to directly edit.]}

\subsection{Assessing Linguistic Knowledge in LMs}
Assessing the linguistic knowledge of LMs has long been a central topic in computational linguistics. 
Early studies focused on assessing the implicit linguistic knowledge of language models emerging from training, such as general syntactic knowledge \cite{gulordava-etal-2018-colorless,goldberg2019assessing,hu-etal-2020-systematic,wilcox-etal-2018-rnn}, dependency structure \citep{hewitt-manning-2019-structural,manning2020emergent}, natural language inference \citep{mccoy-etal-2019-right}, and psycholinguistics judgments \citep{warstadt-etal-2020-blimp-benchmark,ettinger-2020-bert}. 
This line of work centers mostly on English and only measure the implicit linguistic knowledge through proxies like probes, logits, and perplexity.

As LLMs' capacities continue to increase, research has shown that LLMs can follow explicit meta-linguistic concepts or learn languages from explicit meta-linguistic descriptions \cite{tanzer2024a,bean2024lingoly,zhang-etal-2024-hire,spencer-kongborrirak-2025-llms,zhang-etal-2024-teaching,ramos2024grammamt,beguvs2023large}, and can infer the underlying grammatical rules through concrete examples during in-context learning \cite{ginn-etal-2024-teach}. 
Yet, there is still large room for improvement in terms of linguistic reasoning, even for the state-of-the-art LLMs \cite{bean2024lingoly}---most importantly, existing approaches only deal with a handful of low-resource languages and mostly on machine translation tasks. 
It remains unclear if LLMs can perform abstract meta-linguistic reasoning across low-resource languages that are not seen during training. 
In this paper, we evaluate the extent to which LLMs can perform linguistic reasoning across a wide range of structural phenomena and generalize to unseen low-resource languages. 
The ability to make correct inferences would demonstrate the models’ potential to support the analysis of previously understudied languages.

%Building on this line of work, we develop a prompting-based framework that minimizes computational overhead by avoiding fine-tuning and operating at the sentence level. We curate a structured dataset of IGT examples from 18 low-resource and endangered languages, each paired with relevant grammatical descriptions, and design lightweight evaluation tasks that enable systematic assessment of LLMs’ linguistic reasoning across a broad range of typological features.

% SciInstruct: Because reasoning given minimal information is akin to scientific reasoning, fine-tuning on scientific and mathematical reasoning data can also improve model performance.

\section{Data}
\label{sec:data}
We construct our benchmark from a collection of low-resource languages documented in publicly available reference grammars published by \textit{Language Science Press} (LSP),\footnote{\url{https://langsci-press.org/}} an open-access publisher of high-quality linguistic research. 
We select books from their \textit{Studies in Diversity Linguistics} and \textit{Comprehensive Grammar Library} series. 

\subsection{Reference Grammars}
\label{subsec:referenceGrammars}
Reference grammars are comprehensive, systematic descriptions of individual languages, often based on fieldwork and long-term collaboration with native speakers \cite{mosel2006grammaticography,chelliah2013fieldwork}. 
Their goal is to capture linguistic intricacies through various examples and discussions of language use in diverse contexts. 
Typically, they address all major linguistic domains, including phonology, morphology, syntax, semantics, and pragmatics, providing valuable resources for theoretical research, typological comparison, language learning, documentation, and revitalization. 
%The general chapter layout is shown in Figure~\ref{fig:refGrammarStructure}.

For instance, Carol J. Pebley and Thomas E. Payne authored \textit{A Grammar of Kagayanen}: a Western Austronesian language spoken by around 30,000 people in the Philippines \cite{kagayanen}. 
% This grammar is the product of nearly four decades of fieldwork, corpus collection, and community engagement. 
The work adopts a typologically informed descriptive framework inspired by Dixon’s Basic Linguistic Theory \cite{dixon}.

All grammar books used in this study are publicly available under the Creative Commons Attribution 4.0 International License (CC BY 4.0).\footnote{\url{https://creativecommons.org/licenses/by/4.0/} 
This license permits use, distribution, and adaptation of the materials, provided appropriate credit is given to the original authors and source.}

\subsection{Data Preprocessing}
\label{subsec:preprocessing}

\vspace{2pt}\noindent\textbf{Parsing the \LaTeX\ source files.} 
We retrieve the \LaTeX\ source code of 18 reference grammars from their publicly available GitHub sites. 
To ensure the utility of each grammar for our benchmark, we filter each chapter's raw \LaTeX\ source file against our criteria. 
Specifically, we retain languages that (i) include labelled sections (via \texttt{\textbackslash label} tags) that correspond to grammatical rules or descriptive content, and (ii) contain IGT examples that are explicitly linked to these rule explanations. 
For each selected IGT instance, we require that a target keyword (typically a word, morpheme or form under discussion) be highlighted within the example, thereby allowing us to align example sentences with specific grammatical features. 

We begin by converting the raw \LaTeX\ files from each grammar into plain text, removing all formatting commands while retaining boldfaced keywords that indicate grammatical focus. 
Chapters that lack IGT examples, such as acknowledgments and appendices, are excluded in certain languages.

\vspace{2pt}\noindent\textbf{Categorizing individual chapters.} 
%We store our structured data in CSV files, with each row centered around a KP and its associated examples. 
Chapters are manually categorized into either phonology, morphology, syntax, semantics, pragmatics, or other linguistic subfields based on their introductory content (see Appendix~\ref{subsec:chapterCategorization}). 
We exclude chapters related to phonetics (if applicable) due to the lack of IGT content and inconsistent formatting of the symbols from the International Phonetic Alphabet. %The processed dataset is available on our GitHub repository\footnote{\url{https://anonymous.github.io/}}. 

\vspace{2pt}\noindent\textbf{Extracting IGT instances.}  
After cleaning the \LaTeX\ syntax, we extract structured IGT examples from each chapter, along with their preceding \textbf{knowledge points (KPs)}, which we define as explanatory paragraphs containing an IGT label tag and the grammatical rationale for the associated example (see Figure~\ref{fig:predicativeAdjectives}). 
In addition, we record the hierarchical metadata for each example, which includes the chapter title, section heading, and subsection heading.
% We also extract the hierarchical metadata for each example, including chapter title, section and subsection headings, KP text, label, and IGT content.
Each IGT instance is filtered based on structural markers such as label tags, transcription lines (when applicable), morpheme segmentation lines, glossing lines, and free translations. 
Figure~\ref{fig:clean_process} shows an example of Pichi \cite{pichi} from the cleaning process. After this automated extraction, we perform manual cleaning to ensure that all examples have a complete and aligned IGT structure. 
We then verify that the number of words (separated by spaces) matches the number of glosses. In each word, we also ensure one-to-one alignment between morphemes (separated by hyphens) and glosses.

In this study, we use the morpheme-segmented line as standardized input across languages.
% ---since morpheme segmentation from the raw transcription requires additional linguistic knowledge, we avoid adding this extra layer of complexity in order to keep the LLM evaluation focused and interpretable.
% when multiple lines are present, including a raw transcription (without morpheme segmentation) and a morpheme-segmented line, 
This choice reflects finer-grained grammatical units and ensures better alignment with glosses.
% \textcolor{orange}{[Does this mean ``we don't let LLMs do morpheme segmentation whenever possible so that the evaluation is more streamlined'', or ``we don't further split morphemes'' (which I can't make sense of)? I assume it's the former, but what if we don't have the morpheme segmentation? How about removing it?]}
%Because canonical segmentations may differ from surface forms \footnote{Canonical segmentations represent the underlying forms of morphemes, which may differ from how they appear on the surface due to phonological or morphological changes. For example, the word “happiness” is canonically segmented as happy+ness, even though the surface form drops the -y in happy.}, we avoid introducing additional complexity that could confound LLM evaluation.

\begin{figure}[t]
\centering
    \begin{tikzpicture}[font=\sffamily]
  % Top box (no border)
    \fontsize{7.4}{10}\selectfont 
    \node[rectangle, draw=none, fill=white, rounded corners=2pt,
        inner xsep=0pt, inner ysep=0pt, align=left] (before) at (0,0) {
        \verb|  \label{ex:key:127}| \\
        \verb|  \gll Dí  g|\texttt{$\varepsilon$\'{}}\verb|l| \verb| pikín  \textbf{ova}-\textbf{dráy}  ó.\\| \\
        \verb|this girl child over.\textsc{cpd}{}-be.dry \textsc{sp}\\| \\~\\
        \verb|\glt 'This girl is really too lean.' [dj07ae 207]| \\
        \verb|\z|
        };

  % Bottom box (no border)
  \node[rectangle, draw=none, fill=white, rounded corners=2pt,
        inner xsep=10pt, inner ysep=6pt, align=left] (after) at (0,-2.5) {
            \verb|\label{ex:key:127}| \\
            \verb|\gll   Dí    g|\'\textepsilon\verb|l   pikín  \textbf{ova}-\textbf{dráy} ó.| \\
            \verb|\gls   this  girl  child  over.CPD-be.dry            SP| \\
            \verb|\glt   'This girl is really too lean.'|
        };

  % Arrow from top to bottom
  \draw[-Stealth, thick] (before) -- (after);
  \end{tikzpicture}
  \caption{The top portion shows the raw \LaTeX\ source of an IGT example in Pichi \cite{pichi}, where individual morphemes and glosses are annotated using various commands. The bottom portion shows the cleaned version after processing: it converts the gloss line into three aligned components---the morpheme line, gloss line, and translation line.}
  \label{fig:clean_process}
\end{figure}

\section{The \textsc{LingGym} Benchmark}
\label{sec:lingGymDataset}

The high-level characteristics of our \textsc{LingGym} dataset are summarized in Table~\ref{tab:datasetStats}. In total, we process 18 reference grammars from LSP, spanning 8 language families, and yield 19,612 IGT examples aligned with relevant KPs after data filtering and cleaning. 
Most languages in \textsc{LingGym} are from the African and Pacific regions, areas that have traditionally been underrepresented in the NLP community. 
%A summary of their linguistic categories is presented in Table~\ref{tab:subfieldDistribution}. 
A summary of the dataset's distribution across linguistic subfields is shown in Table~\ref{tab:subfieldDistribution}: the benchmark covers all aspects of the linguistic subfields commonly used to describe the structures of languages, with a strong focus on syntax in the questions reflecting the typical emphasis in language documentation practices.

% \definecolor{AC}{RGB}{204,230,255}  
\definecolor{AC}{RGB}{51,34,136}
% \definecolor{AU}{RGB}{204,255,204}   
\definecolor{AU}{RGB}{17,119,51}
% \definecolor{TNG}{RGB}{255,229,204}  
\definecolor{TNG}{RGB}{68,170,153}
% \definecolor{IE}{RGB}{255,255,204} 
\definecolor{IE}{RGB}{136,204,238}
% \definecolor{ST}{RGB}{255,204,229}  
\definecolor{ST}{RGB}{221,204,119}
% \definecolor{QU}{RGB}{229,204,255}   
\definecolor{QU}{RGB}{204,102,119}
% \definecolor{NE}{RGB}{224,224,224}
\definecolor{NE}{RGB}{170,68,153}
% \definecolor{NS}{RGB}{255,204,204}   
\definecolor{NS}{RGB}{136,34,85}

\begin{table}[ht]
  \centering
  \small
  \begin{tabular}{llr}
    \toprule
    \textbf{Language}       & \textbf{Family}            & \textbf{Examples} \\
    \midrule
    \rowcolor{AC!35} Pichi          & Atlantic-Congo          & 2,846 \\
    \rowcolor{AC!35} Gyeli          & Atlantic-Congo          &   691 \\
    \rowcolor{AC!35} Moloko         & Atlantic-Congo          &   439 \\
    \rowcolor{AC!35} Fwe            & Atlantic-Congo          &   147 \\
    \rowcolor{AU!35} Papuan Malay   & Austronesian            & 3,766 \\
    \rowcolor{AU!35} Rapa Nui       & Austronesian            & 1,709 \\
    \rowcolor{AU!35} Kagayanen      & Austronesian            &   550 \\
    \rowcolor{AU!35} Vamale         & Austronesian            &    67 \\
    \rowcolor{TNG!35} Komnzo        & Trans-New Guinea        &   709 \\
    \rowcolor{TNG!35} Mauwake       & Trans-New Guinea        & 1,787 \\
    \rowcolor{TNG!35} Kalamang      & Trans-New Guinea        &   656 \\
    \rowcolor{TNG!35} Ulwa          & Trans-New Guinea        & 1,851 \\
    \rowcolor{IE!35} Palula         & Indo-European           & 1,674 \\
    \rowcolor{IE!35} Tuatschin      & Indo-European           & 1,113 \\
    \rowcolor{ST!35} Japhug         & Sino-Tibetan            &   358 \\
    \rowcolor{QU!35} Yauyos Quechua & Quechuan                & 1,143 \\
    \rowcolor{NE!35} Mehweb         & Northeast Caucasian     &    85 \\
    \rowcolor{NS!35} Ik             & Nilo-Saharan            &    21 \\
    \midrule
    \multicolumn{2}{r}{\textbf{Total}}                          & \textbf{19,612} \\
    \bottomrule
  \end{tabular}
  \caption{Number of KP-IGT pairs for the  \textsc{LingGym} dataset. In total, 18 reference grammars from 8 language families are processed.} 
  \label{tab:datasetStats}
\end{table}

\begin{table}[ht]
  \centering
  \small
  \begin{tabular}{@{}lrr@{}}
    \toprule
    \textbf{Linguistic Subfield} & \textbf{\# Examples} & \textbf{\% of Total} \\
    \midrule
    Morphology   & 1,410 & 7.19\% \\
    Phonology    &    71  & 0.36\% \\
    Pragmatics   &   139  & 0.71\% \\
    Semantics    &   967  & 4.93\% \\
    Syntax       &16,747 & 85.39\% \\
    Other        &   278  & 1.42\% \\
    \midrule
    \textbf{Total} & \textbf{19,612} & \textbf{100\%} \\
    \bottomrule
  \end{tabular}
  \caption{Distribution of examples by linguistic subfield.}
  \label{tab:subfieldDistribution}
\end{table}

\subsection{Word-Gloss Inference}
\label{subsec:morphemeGlossInference}
We introduce a multiple-choice, cloze-style word/word-gloss inference task, which can be used to evaluate whether LLMs can infer grammatical information from structured linguistic data. 
Each question presents an IGT example in which a single word or a single word plus its gloss has been masked.
The model must identify the correct word or word-gloss pair from four options, based on the sentence context, grammatical structure, and accompanying explanation. 
More details of the task can be found in \S\ref{subsec:difficultyLevels}.

\subsection{Question Generation}
\label{subsec:questionGeneration}
We generate these questions using examples drawn from our cleaned data.
For each instance, if a word or any of its morphemes is marked with a \texttt{\textbackslash textbf} tag, we identify that word as the target. 
To create the set of four answer choices (one correct answer and three distractors), we employ three strategies to generate plausible distractors: 
\begin{itemize}[leftmargin=*,itemsep=0pt,topsep=2pt]
    \item \textbf{Form-based distractor (LCS-based)}: 
    We find a distractor gloss that shares the longest common substring (LCS) with the correct gloss but differs in grammatical function. 
    For example, given the correct word-gloss pair walk--PST (walk-ed), we generate a distractor like walk--PROG (walk-ing). This shares the root \textit{walk} (via the LCS) but fulfills a different grammatical function. 
    \item \textbf{Semantics-based distractor}: 
    We compute the semantic similarity between glosses by embedding them using Sentence-BERT \cite{reimers-gurevych-2019-sentence}. 
    The gloss that has the highest semantic similarity with, but is not identical to, the correct gloss is selected and mapped back to the corresponding word in the dataset. 
    This approach introduces subtle meaning contrasts to test deeper grammatical understanding. 
    \item \textbf{Chapter-local distractor}: 
    To promote lexical and structural diversity, we randomly sample a word-gloss pair from the same grammar chapter, ensuring that the distractor does not overlap in form or gloss with any of the other options. 
    This approach adds noise that reflects the topic domain but avoids trivial elimination. 
\end{itemize}
All distractor candidates are also ensured not to overlap with each other. 
To prevent positional bias in candidate answers, we randomly assign the correct answer to one of the four choice positions in each question. 
This randomization is applied uniformly across all examples, ensuring that each position (A--D) contains the correct answer approximately 25\% of the time.

To construct each question, we mask all correct choice words in the gloss and knowledge point lines. 
The masking in the surface line is always performed at the word level, ensuring consistent granularity across examples. 
This masking approach preserves the context while clearly signalling the missing element to the model. 
However, masking in the free translation line presents a challenge, as translations often paraphrase or use semantically related expressions rather than a direct lexical equivalent of the source word/morpheme. 
As a result, the corresponding segment in the translation cannot always be reliably identified or removed without altering the naturalness or interpretability of the sentence---this introduces a limitation in our masking approach: while the surface and gloss lines are systematically masked, the translation may still contain indirect cues about the target word.

% \texttt{RegEx} is used to extract the model's translation, and is standardized by removing ASCII punctuation and making all characters lowercase. The same procedure is applied to the reference.

\begin{figure}[ht]
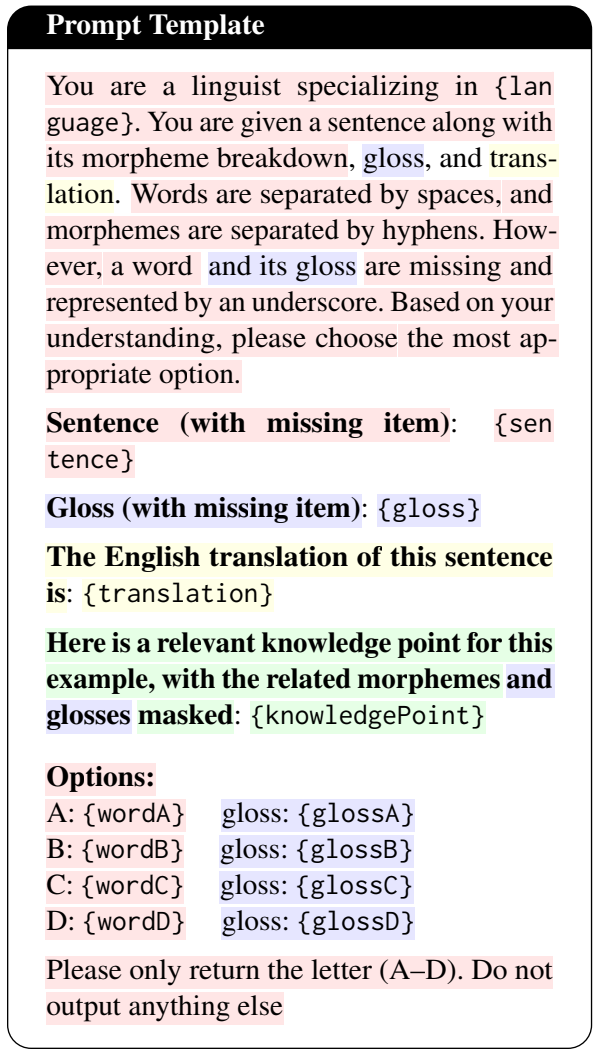

        \begin{tcolorbox}[
        enhanced,
        colback=white,
        colframe=black,
        arc=3mm,
        boxrule=0.5pt,
        title=\textbf{Prompt Template}
        ]
        \SHL{You are a linguist specializing in \texttt{\{language\}}.  You are given a sentence along with its morpheme breakdown}, 
        \GHL{gloss}, and 
        \THL{translation}. 
        \SHL{Words are separated by spaces, and morphemes are separated by hyphens. However, a word } 
        \GHL{and its gloss} 
        \SHL{are missing and represented by an underscore. Based on your understanding, please choose the most appropriate option.}
        
        \vspace{1ex}
        \textbf{\SHL{Sentence (with missing item)}}: \SHL{\texttt{\{sentence\}}}
        
        \vspace{1ex}
        \textbf{\GHL{Gloss (with missing item)}}: \GHL{\texttt{\{gloss\}}}
        
        \vspace{1ex}
        \textbf{\THL{The English translation of this sentence is}}: \THL{\texttt{\{translation\}}}
        
        \vspace{1ex}
        \textbf{\KPHL{Here is a relevant knowledge point for this example, with the related morphemes}
        \GHL{and glosses} \KPHL{masked}}: \KPHL{\texttt{\{knowledgePoint\}}}
        
        \vspace{2ex}
        
        \SHL{\textbf{Options:}}\\
        \SHL{A: \texttt{\{wordA\}}}  \quad \GHL{gloss: \texttt{\{glossA\}}}\\
        \SHL{B: \texttt{\{wordB\}}}  \quad \GHL{gloss: \texttt{\{glossB\}}}\\
        \SHL{C: \texttt{\{wordC\}}}  \quad \GHL{gloss: \texttt{\{glossC\}}}\\
        \SHL{D: \texttt{\{wordD\}}}  \quad \GHL{gloss: \texttt{\{glossD\}}}
        
        \vspace{1ex}
        
        \SHL{Please only return the letter (A–D). Do not output anything else}
        \end{tcolorbox}
    \caption{The prompt template used across different difficulty levels.}
    \label{fig:fullPromptTemplate}
\end{figure}

\subsection{Difficulty Levels}
\label{subsec:difficultyLevels}
To evaluate the impact of different types of linguistic information, we design our prompts to include the following types of knowledge:
\begin{itemize}[leftmargin=*,topsep=2pt,itemsep=0pt]
    \item \SHL{Original sentence (S)}: the morpheme-segmented sentence in target language.
    \item \GHL{Gloss information (G)}: The glosses for the given words.
    \item \KPHL{Knowledge points (KP)}: The relevant knowledge points in the grammar book.
    \item \THL{English translations (T)}: The English translation.
\end{itemize}
We conduct our main experiments with four difficulty levels based on data availability: \textbf{S}, \textbf{S+G}, \textbf{S+G+KP}, and \textbf{S+G+KP+T}. 
All prompts follow the template displayed in Figure~\ref{fig:fullPromptTemplate}, and an example is shown in Figure~\ref{fig:pipeline}. 

\begin{table*}[ht]
\centering
\small
%\begin{scriptsize}
\begin{tabular}{@{\extracolsep{\fill}}l*{2}{S}*{3}{S}*{2}{S}*{2}{S}S}
\toprule
&
\multicolumn{2}{c}{\textbf{Qwen2.5}} &
\multicolumn{3}{c}{\textbf{Gemma 3}} &
\multicolumn{2}{c}{\textbf{DeepSeek‑R1}} &
\multicolumn{2}{c}{\textbf{LLaMA3}} &
\multicolumn{1}{c}{\textbf{GPT-4}} \\
\cmidrule(lr){2-3} \cmidrule(lr){4-6} \cmidrule(lr){7-8} \cmidrule(lr){9-10} \cmidrule(lr){11-11}
 Difficulty & {7B} & {32B} & {4B} & {12B} & {27B} & {7B} & {32B} & {8B} & {70B} & {o4-mini} \\
\midrule
S             & 33.04 & 38.66 & 32.74 & \textbf{43.63} & 41.48  & 33.39  &39.62  & 29.62 &34.46 & 41.74 \\
S+G           & 41.64 & 46.75 & 38.88 & 47.03 & \textbf{48.17}  & 35.24  & 48.16  & 30.83 &42.37 & 46.02 \\
S+G+KP         & 56.08 & 60.97 & 49.76 & 59.47 & 61.83  & 46.18  &\textbf{65.50}  & 39.44 &59.64 & 57.28 \\
S+G+KP+T       & 71.09 & 78.29 & 63.92 & 73.97 & 77.02  & 54.39  &\textbf{81.17}  & 50.32 &78.25 & 73.88 \\
\bottomrule
\end{tabular}
\caption{Accuracies for all languages across input settings and models.}
\label{tab:acc-main-results}
%\end{scriptsize}
\end{table*}

\begin{figure*}[t]
\centering
  \includegraphics[width=0.8\textwidth]{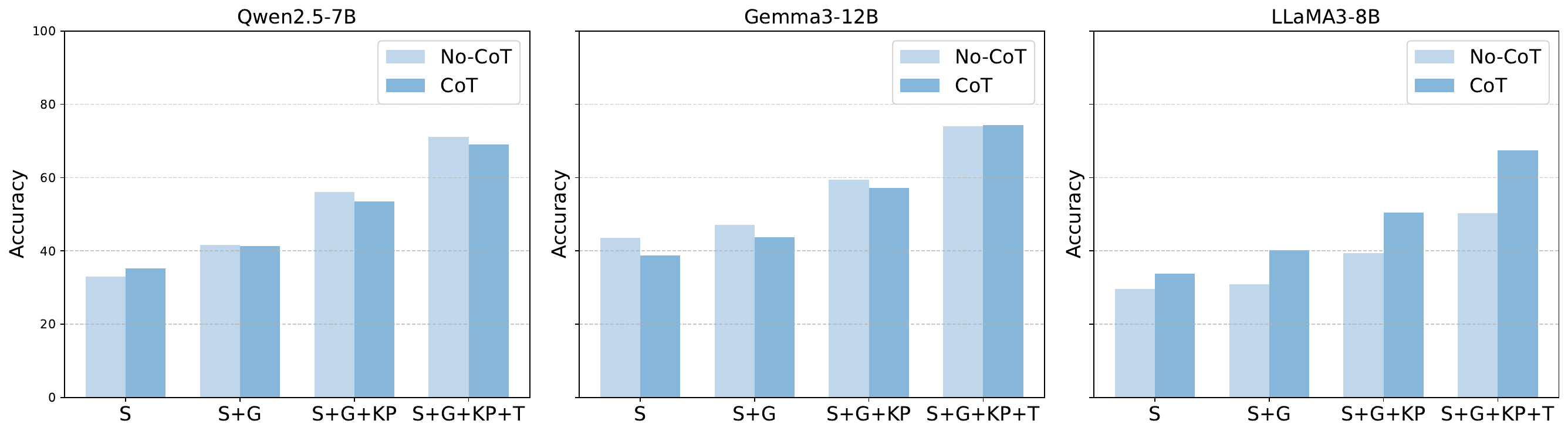}
  \caption{Accuracies of CoT vs. No-CoT prompting across different models and input settings.}
  \label{fig:CoT}
\end{figure*}

\section{Experiments and Results}
\label{sec:experimentsAndResults}
\subsection{Experimental Setup}
\label{subsec:experiments}
We evaluate a diverse set of publicly available LLMs, covering a range of sizes and model families. 
Our evaluation includes models from four major families: \textbf{Qwen2.5} \cite{qwen2.5}, \textbf{Gemma 3} \cite{team2025gemma}, \textbf{DeepSeek-R1} \cite{guo2025deepseek}, and \textbf{LLaMA3} \cite{dubey2024LLaMA}. 
For Qwen2.5, we include the 7B and 32B models; for Gemma3, we evaluate the 4B, 12B, and 27B variants; for DeepSeek-R1, we test the 7B and 32B; and for LLaMA3, we assess the 8B and AWQ-quantized 70B models. 
We use the AWQ quantization \cite{lin2024awq} for larger 70B models due to limited computing resources. 
All models are instruction-tuned and are accessed via open-source platforms (i.e., HuggingFace Hub). 
Inference was performed through \texttt{vLLM} \cite{kwon2023efficient} and \texttt{transformers} \cite{wolf-etal-2020-transformers}. 
All experiments were run on A6000 Ada GPUs, with more details provided in Appendix \ref{sampling}. 
For evaluation, since the word-gloss inference task is formulated as multiple-choice questions with balanced choice distributions, we report standard accuracy as the primary evaluation metric.

\begin{table}
\begin{adjustbox}{width=\columnwidth,center}
\centering
\begin{tabular}{lcccc}
\toprule
\textbf{Difficulty} & \textbf{Qwen2.5-7B} & \textbf{Gemma3-12B} &  \textbf{DeepSeek-R1-7B} &\textbf{LLaMA3-8B} \\
\midrule
S          &  33.04    &  43.63     & 33.39 & 29.62     \\\midrule
S+G          & 41.64      & 47.03   & 35.24  &  30.83     \\
S+T         &  48.65     & 68.52   & 53.91 &  48.65     \\
S+KP         &  52.78     & 55.76   & 45.22  & 46.22    \\\midrule
S+G+KP       &  56.08     &  59.47   &46.18  &  39.44    \\
S+G+T       & 66.59   &  69.43  &  54.39 & 54.39     \\
S+KP+T       &  59.12    & 72.95    & 58.79 &  59.12   \\
S+G+KP+T     &  71.09     &  73.97   & 54.39   & 50.32      \\
\bottomrule
\end{tabular}
\end{adjustbox}
\caption{Accuracies across all information permutations for selected models. Full results are shown in Appendix~\ref{sec:cot-full}.}
\label{tab:accuracyAllDifficulties}
\end{table}

\begin{figure*}[t]
\centering
  \includegraphics[width=0.9\textwidth]{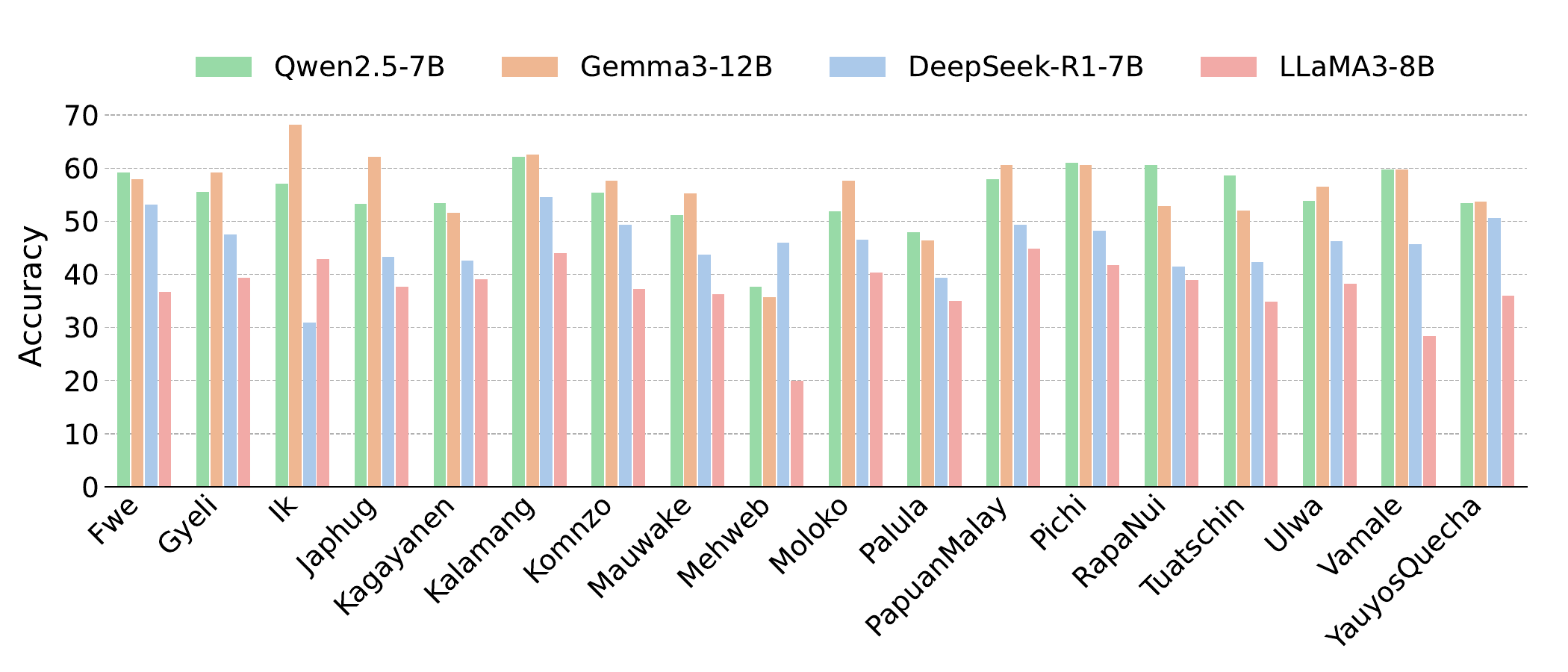}
  \caption{Weighted average accuracy scores across languages under the S+G+KP setting for select models.}
  \label{fig:langFamilyScores}
\end{figure*}

\subsection{Results} 
%\freda{Do we really want to have a separate section?}
\label{subsec:results}
Our main results (Table~\ref{tab:acc-main-results}) present accuracies for all evaluated LLMs across four difficulty levels. 
More detailed results are provided in Appendix~\ref{sec:fullResults}, with a concrete prompt example (S+G+KP+T) and model prediction results shown in Figure \ref{fig:examplefullPromptTemplate}.

\vspace{2pt}\noindent\textbf{The meta-linguistic reasoning benchmark is challenging to LLMs despite data contamination issues.} 
Data contamination is a common issue in many LLM benchmarks \cite{sainz-etal-2023-nlp,deng-etal-2024-investigating}, as LLMs are trained on almost all found data on the Internet. 
All reference grammar books we processed are subject to this risk, as they are openly accessible as \LaTeX\ source code hosted on GitHub. 
To clarify the potential impact of data contamination, we test the LLM's performance only with the raw sentences in evaluation languages, without providing any additional information---if LLMs perform above the chance level (25\%), it is likely that they have seen some of the language during pretraining.

Indeed, we find evidence of potential data contamination (first row in Table~\ref{tab:acc-main-results}): all LLMs have above-chance performance even when provided only the original sentences. 
Larger models tend to memorize even more, evidenced by higher performance.
However, the overall performance is still far from perfect, suggesting that the memorization effect is limited; that is, our dataset serves as a meaningfully challenging benchmark in a highly specialized domain.

\begin{figure*}[t]
\centering
  \includegraphics[width=\textwidth]{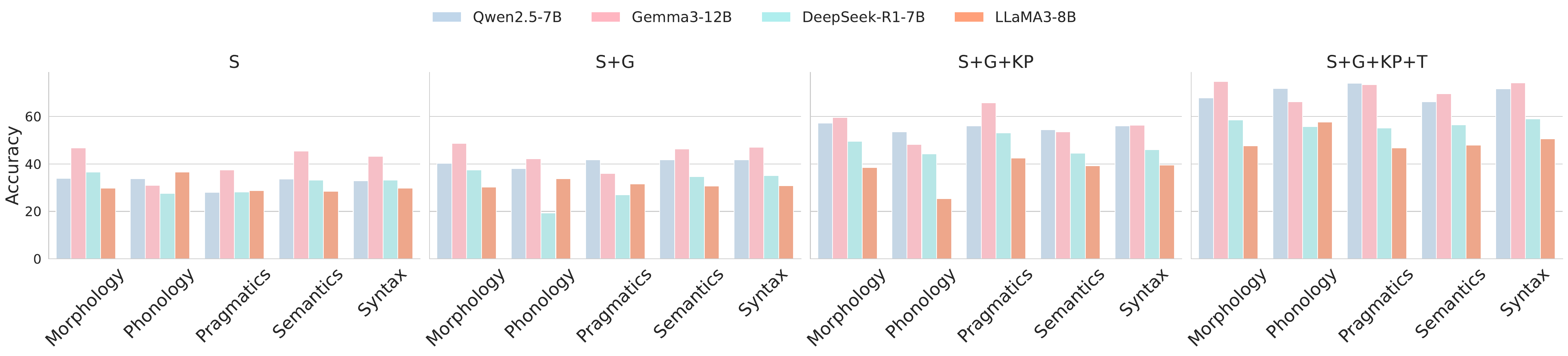}
  \caption{Weighted average accuracies of selected language models across five linguistic subfields: morphology, phonology, pragmatics, semantics, and syntax, under four levels of input difficulty.}
  \label{fig:subfieldScores}
\end{figure*}

\vspace{2pt}\noindent\textbf{KPs improves performance across all conditions.}
In line with earlier work \cite{tanzer2024a,zhang-etal-2024-hire}, we find that adding KPs brings consistent improvements across LLM families and parameters (Table~\ref{tab:acc-main-results}), suggesting that LLMs possess some abilities to comprehend the linguistic concepts in KPs and associate them with concrete language examples. 
As expected, larger models outperform smaller models by a large margin. 
The best performing model, DeepSeek-R1 32B, reaches around 81\% accuracy, suggesting that LLMs show remarkable capabilities in meta-lingusitic reasoning that is independent of languages. 

To further validate this effect, we conduct controlled experiments on selected models by testing LLMs across all difficulty condition permutations. 
Our ablation results from Table~\ref{tab:accuracyAllDifficulties} indicate that gloss information, English translations, and knowledge points each contribute to the meta-linguistic reasoning, independent of each other. Yet none of the LLMs achieve perfect accuracy on these tasks, suggesting a large room for improvement.

\vspace{2pt}\noindent\textbf{CoT does not bring clear improvement to the performance.} 
Chain-of-Thought (CoT) prompting \cite{wei2022chain} has been shown to effectively improve performance on reasoning tasks, although the improvement is mainly limited to math and symbolic reasoning tasks \cite{sprague2025to}. 
As shown in Figure~\ref{fig:CoT}, we do not find conclusive evidence that meta-linguistic reasoning benefits much from CoT across all LLMs from different families. 

Reasoning models like DeepSeek-R1 and o4-mini are also not competitive with non-reasoning models. 
The only exception is DeepSeek-R1 32B \cite{guo2025deepseek}, a reasoning model trained to perform long CoT. 
Although DeepSeek-R1 32B dominates in almost all conditions, DeepSeek-R1 7B does not exhibit such an advantage.

\vspace{2pt}\noindent\textbf{LLM performance is relatively similar across individual languages, language families, and linguistic subfields.} The full table by language and models can be found in Appendix~\ref{sec:lingSubfieldResults}. Figure~\ref{fig:langFamilyScores} indicates that the performance is relatively stable across benchmarks, despite some minor variations. This further validates the efficacy of our benchmark, indicating that our benchmark is representative and balanced within and across each language. 

As \textsc{LingGym} is sourced from the whole reference grammar books, it covers structural descriptions in all linguistic subfields that are considered necessary to describe a language. 
As shown in Figure~\ref{fig:subfieldScores}, performance does not vary substantially across linguistic subfields, aside from minor variations. 
This suggests that LLMs are able to reason---at least to some extent---across linguistic subfields represented in these reference grammars. 

\subsection{Error Analysis}
\label{subsec:errorAnalysis}
We further investigate errors made under our strongest configuration, DeepSeek-R1 32B, with sentence, gloss, knowledge points, plus translation lines (S+G+KP+T) based on the model’s predictions. Most failures can be categorized into three main types:

\vspace{2pt}\noindent\textbf{Abbreviation-heavy items with opaque gloss tags.}
When correctness depends on understanding dense sequences of gloss abbreviations (such as \texttt{PP4-CON=DEM.I7}\footnote{This gloss is extracted from Fwe. \texttt{PP4} = pronominal prefix with agreement set 4; \texttt{CON} = connective; \texttt{DEM.I7} = demonstrative (series I, form 7; grammar-specific).} and \texttt{OBJ.LINK-PL}\footnote{This gloss is from Gyeli. \texttt{OBJ.LINK} = object linking H tone; \texttt{PL} = plural marker.}), the model appears to treat the tags as uninterpreted symbols and guess among look-alike forms. For example, in a Moloko sentence where the prompt gloss already encodes number and possession (``\texttt{children=Pl\footnote{\texttt{Pl} = plural noun clitic.} \underline{\hspace{1cm}} POSS=1S.POSS=Pl}\footnote{\texttt{POSS} = possessive pronoun; “.” stacks features inside one tag; \texttt{1S} = first person singular.}'', with the translation ``These particular children here belong to me.''), the model prefers \texttt{DEM=Pl}\footnote{\texttt{DEM} = demonstrative.} over the correct bare \texttt{DEM}. Similarly, in Komnzo for the phrase (``Do it here / with the tools here.''), the model chooses the bare \textit{zane} (glossed as \texttt{DEM:PROX}\footnote{\texttt{PROX} = proximal demonstrative. }), whereas the correct answer is \textit{zane=me} (glossed as \texttt{DEM:PROX=INS}\footnote{\texttt{INS} = instrumental case.}). The instrumental clitic \texttt{=me}  is the decisive element that the model fails to recognize. This indicates that comprehensive explanations of these abbreviations need to be incorporated into the prompts as well. We will treat this enhancement as future work.

\vspace{2pt}\noindent\textbf{Semantically similar distractors.}
The model often selects an option that is plausible in English but morphosyntactically ill-formed in the target language. In Kalamang, for the sentence ``I particularly like doing that thing.'', the model chooses \textit{great} over the gold \textit{gladly}. Both convey positive affect in English, but only \textit{gladly} fits the required collocational/morphological slot. A similar error occurs in Moloko: given the gloss ``\texttt{1S+IFV-see\footnote{\texttt{1S} = first person singular; \texttt{IFV} = imperfective aspect.} goat=1S.POSS=Pl\footnote{\texttt{POSS} = possessive; \texttt{Pl} = plural noun clitic.} three 2S\footnote{\texttt{2S} = Second person singular.} \underline{\hspace{1cm}}}'' with its translation ``I see my three goats that you gave to me'', the model predicts \textit{am\textschwa{}-v\textschwa{}l=\textopeno{}k\textsuperscript{w}} (glossed as \texttt{DEP-give=2S.IO}\footnote{\texttt{DEP} = dependent form of the verb; \texttt{2S.IO} = 2nd-person singular indirect object pronominal.}), whereas the correct form is \textit{am\textschwa{}-v\textschwa{}l=aw} (glossed as \texttt{DEP-give=1S.IO}\footnote{\texttt{1S.IO} = first person singular indirect object pronominal.}). This suggests that the model struggles to differentiate between synonyms or semantically related terms when precise morphological constraints are involved. The underlying issue appears to be that the model relies on semantic similarity rather than understanding the specific grammatical requirements of the target language structure. 

\vspace{2pt}\noindent\textbf{Fine-grained form differences (tones, vowels).}
The model also struggles when answer choices differ only by minimal morpho-orthographic features such as tone marks or single vowels. 
In Ulwa, when selecting a word meaning ``also'', the model chooses \textit{maweka} despite the correct form being \textit{moweka}. 
Both forms have the same meaning, but the single vowel difference completely changes the correctness of the answer. 
This indicates that the model has insufficient knowledge of phonological and orthographic variants, resulting in treating morphologically distinct forms as interchangeable alternatives.

\section{Conclusion}
We present \textsc{LingGym}, a comprehensive benchmark to assess the meta-lingusitic reasoning ability of LLMs in 18 languages across all linguistic aspects. 
Our analyses show that LLMs exhibit some capabilities to perform meta-linguistic reasoning, %as their task performance benefits from the explicit linguistic knowledge in reference grammars. These results 
highlighting the potential of using LLMs to assist linguistic analysis.

Our benchmark emphasizes mapping abstract linguistic rules to concrete sentences. 
Yet in actual fieldwork, it is also important to induce linguistic rules from linguistic samples, which might be assisted with LLMs \cite{spencer-kongborrirak-2025-llms}. 
In the future, we will extend our work to cover more diverse and in-depth use cases for linguistic analysis, especially for low-resource and endangered languages.

%We present a high-quality, cleaned dataset of IGT examples from 18 endangered low-resource languages documented in \textit{Language Science Press}. The structured CSV files include not only KPs but also the parent section and subsection paragraphs that allow researchers to control the amount of language-specific information contained in input prompts. Preliminary knowledge-augmented prompting results on the Pichi language show that including both high-level features (e.g., word order) and finer-grained details helps LLMs output translations that better correspond with human judgments (c.f. \texttt{chrF} score), even with minimal model requirements. We hope this resource facilitates future work in enhancing LLM reasoning in the realm of NMT for low-resource languages and serves as a stepping stone towards automated language documentation.

\section*{Limitations}
%Even though there are writing conventions for books published in \textit{Language Science Press}, glossing conventions are not standardized. Even within the Leipzig glossing abbreviations \cite{leipzig_glossing_rules} in IGT data, each book's authors may define unique LaTeX commands or have varying capitalization patterns. We check for validity only by ensuring the number of morphemes and glosses match, but these discrepancies may distract the model.

%Furthermore, many rarer characters cannot be expressed as Unicode in \texttt{.tex} files, so several authors use unique commands to render them. While there are resources that list commands to render upwards of 5000 symbols \cite{pakin2008comprehensive}, we stick with the thorough (but arguably naive) RegEx cleaning method we described in \ref{subsec:preprocessing}.

Linguistic analysis is inherently theory-laden and value-laden \cite{bird2020decolonising}. Our benchmark is still limited in scope. The grammatical analyses from most reference grammars follow the structuralist framework, which is only one of the many theoretical frameworks in linguistics. 

Linguistic analysis is a complex task. 
The immediately preceding KP often does not paint the full picture of a given grammatical construction (i.e., extracted KPs often make references to parent subsections or sections), though they still constitute a good starting point. 

Our study only analyzes 18 languages. 
While these languages are understudied within the NLP community, they only represent a tiny fraction of human languages. 
Grambank, a linguistic typological database, records reference grammar books or papers for around 2400 languages \cite{skirgardGrambankRevealsImportance2023}, and we will continue to expand our analyses to more languages. 

Our dataset is imbalanced across linguistic subfields, reflecting the natural skew of reference grammars, which devote disproportionate attention to syntax. 
In principle, subdividing the syntax bucket into finer categories (e.g., word/constituent order, agreement, clause structure) would yield more diagnostic analyses. 
In practice, however, the specific syntactic topics covered vary substantially across languages and sources, which makes a uniform subcategorization scheme difficult to apply consistently and limits cross-language comparability. 
We therefore report results at a coarse level---extending to stable and finer-grained syntactic categories is a valuable direction for our future work.

While we have attempted to evaluate LLMs across model families and parameter counts, due to limited budget, we were not able to evaluate on the larger state-of-the-art models like DeepSeek-R1-671B, o4, and Gemini 2.5 Pro. These models might demonstrate stronger abilities than the models reported. 

\section*{Ethics Statement}

We only selected the reference grammar books that are publicly available with permissive Creative Commons licenses, allowing us to reprocess and redistribute the dataset. 

Our study falls into the scope of fundamental research in natural language processing and linguistics, with the goal of assisting language documentation with LLMs. There is no direct harm associated with this type of research. We expect this work to contribute to the analysis and documentation of endangered languages. 

\ifreview
\else
\section*{Acknowledgements}
We thank three anonymous reviewers and the area chairs for their thoughtful comments on the original manuscript. We would also like to extend our gratitude to the field workers and the community members documenting these languages. This research was enabled in part through the computational resources provided by Advanced Research Computing at the University of British Columbia and the Digital Research Alliance of Canada. The research activities were also supported by the NSERC Discovery Grant and the CFI-JELF Grant awarded to JZ, and a Canada CIFAR AI Chair Award to FS.
\fi 

\bibliography{anthology,custom}
%\bibliography{custom}

\clearpage
\appendix

\clearpage
\section{Sampling Parameters of LLMs}
\label{sampling}
\begin{table}[h]
\begin{adjustbox}{width=\columnwidth,center}
\begin{tabular}{ll}
\toprule
\textbf{Parameter} & \textbf{Value / Description} \\
\midrule
Temperature & 0.7 \\
Top-$p$ & 0.9 \\
Max Tokens & 2048 \\
Repetition Penalty & 1.1 \\
Decoding Strategy & Sampling-based decoding \\
\bottomrule
\end{tabular}
\end{adjustbox}
\caption{Sampling parameters used for LLM generation.}
\label{tab:sampling-params}
\end{table}

\clearpage
\onecolumn
\section{Full Results}
\label{sec:fullResults}

\begin{scriptsize}
\begin{longtable}{
  @{\extracolsep{\fill}}
  >{\scriptsize}l  
  l                
  *{6}{c} ll *{3}{c}  % Total 12 columns
}
\toprule
 &  & \multicolumn{2}{c}{\textbf{Qwen2.5}} &
      \multicolumn{3}{c}{\textbf{Gemma 3}} &
      \multicolumn{2}{c}{\textbf{DeepSeek-R1}} &
      \multicolumn{2}{c}{\textbf{LLaMA3}} &
      \multicolumn{1}{c}{\textbf{GPT-4}} \\
\cmidrule(lr){3-4} 
\cmidrule(lr){5-7}  
\cmidrule(lr){8-9}   
\cmidrule(lr){10-11} 
\cmidrule(lr){12-12}
\textbf{Language} & \textbf{Difficulty}
         & {7B} & {32B}
         & {4B} & {12B} & {27B}
         & {7B} & {32B}
         & {8B} & {70B}
         & {o4-mini} \\
\midrule
\multicolumn{12}{l}{\textbf{Atlantic-Congo Language Family}} \\
Pichi & S & 31.48  & 37.91  & 33.45 &44.27  &42.83   & 30.97  & 41.43   &30.29  &36.33 &42.45\\
&   S+G & 46.03  &  48.31  & 43.57 &49.51  &51.05   & 37.15  & 53.03   &35.10  & 48.91 &47.79 \\
& S+G+KP& 61.00  &  65.32  & 57.34 &62.02  &65.43   & 48.25  & 68.75   &41.78  & 66.90 & 60.40 \\
 & S+G+KP+T& 73.86  &  79.86  & 69.15 &75.86  &79.23   & 60.26  & 81.24   &53.51  & 82.31 & 75.47\\

 Gyeli & S & 26.48   &  29.52  & 30.10 &34.44  &30.68   & 32.88  & 27.49   &30.68  & 28.51 &32.13 \\
&  S+G & 35.75  &  39.13  & 35.31 &39.94  &39.94   & 34.99  & 38.14   &20.09  & 34.15 &40.09 \\
 & S+G+KP& 55.57  &  60.78  & 46.89 &60.35  &63.53   & 47.54  & 63.50   &39.36  & 58.61 & 57.60\\
 & S+G+KP+T & 64.25  &  72.21  & 56.44 &67.29  &73.95   & 57.16  & 76.06   &47.03  & 71.92 &70.62 \\

Moloko  &S & 25.97  &  29.84  & 28.93 &41.46  &33.26   & 29.76  & 28.34   &22.78  & 27.56 &34.40 \\
 &  S+G & 38.29  &  36.90  & 33.71 &46.47  &47.38   & 30.68  & 36.83   &28.70  & 35.31 &43.51 \\
&  S+G+KP& 51.94  &  61.05  & 47.84 &59.68  &63.55   & 46.42  & 64.45   &40.32  & 64.01 &59.45 \\
& S+G+KP+T& 67.88  &  77.90  & 65.15 &71.75  &78.36   & 57.14  & 80.59   &45.79  & 81.55 & 72.89 \\

 Fwe & S & 35.37  &  40.82  & 37.41 &42.86  &36.73   & 29.55  & 33.33   &29.93  & 38.78 &36.73\\
& S+G & 44.22  &  47.62  & 43.54 &40.14  &40.82   & 32.35  & 50.00   &29.25  & 34.01 &41.50\\
 & S+G+KP& 59.18  &  63.27  & 49.66 &57.14  &63.27   & 53.44  & 69.44   &36.73  & 61.09 &63.27  \\
 & S+G+KP+T & 65.31  &  77.55  & 63.27 &69.39  &73.47   & 55.47  & 78.32   &48.98  & 71.43 & 68.71\\
\hline
\multicolumn{12}{l}{\textbf{Austronesian Language Family}} \\
Papuan Malay & S & 39.80  &  49.76  & 42.30 &50.82  &49.52   & 38.37  & 49.20   &33.30  & 42.14 &54.30 \\
& S+G & 43.12  &  51.73  & 43.63 &52.89  &50.77   & 38.46  & 52.08   &32.58  & 45.42 &54.12\\
 & S+G+KP& 57.89  &  63.52  & 52.52 &63.78  &65.37   & 48.28  & 68.30   &44.80  & 62.42 &62.83 \\
&  S+G+KP+T & 77.51  &  82.97  & 68.91 &80.64  &81.41   & 64.38  & 84.73   &60.12  & 81.86 &81.12 \\

 Rapa Nui  & S& 38.68  &  37.62  & 28.91 &43.89  &40.26   & 28.38  & 43.10   &31.01  & 40.78&39.56  \\
&  S+G & 51.84  &  53.00  & 41.95 &46.58  &48.51   & 35.19  & 53.57   &34.87  & 53.31 &52.08 \\
& S+G+KP& 60.62  &  63.39  & 46.99 &57.28  &55.59   & 41.53  & 65.62   &38.99  & 64.97&53.01  \\
& S+G+KP+T& 73.96  &  82.16  & 62.79 &70.39  &72.62   & 52.59  & 81.98   &47.51  & 80.68 &70.39 \\

Kagayanen & S & 30.36  &  38.18  & 33.09 &39.82  &39.82   & 30.26  & 38.64   &25.64  & 34.73 &40.91  \\
& S+G& 43.64  &  46.73  & 39.82 &41.82  &45.82   & 32.45  & 48.70   &32.18  & 47.09 &45.45 \\
& S+G+KP& 53.45  &  55.64  & 50.73 &53.64  &57.45   & 42.70  & 65.06   &39.09  & 56.00 &54.18 \\
&  S+G+KP+T & 66.73  &  72.41  & 61.64 &69.82  &72.36   & 55.70  & 80.22   &52.73  & 79.05 &71.27 \\

Vamale & S & 31.34  &  29.85  & 35.82 &43.28  &32.84   & 39.06  & 38.81   &31.34  & 35.82 &38.81  \\
 & S+G& 34.33  &  47.76  & 44.78 &50.57  &38.81   & 38.46  & 48.44   &23.88  & 29.85 &43.28  \\
 & S+G+KP& 59.70  &  55.22  & 56.72 &50.75  &56.72   & 45.31  & 67.16   &28.36  & 46.27 &55.22 \\
 & S+G+KP+T& 68.66  &  80.60  & 74.63 &71.64  &77.61   & 63.64  & 82.09   &50.75  & 74.63 &73.13 \\\hline

\multicolumn{12}{l}{\textbf{Trans-New Guinea Language Family}} \\
Komnzo  & S & 34.41  &  34.27  & 31.88 &41.47  &42.88   & 32.68  & 31.23   &26.09  & 27.22 &33.99 \\
 & S+G & 43.16  &  42.74  & 39.63 &43.86  &50.49   & 35.61  & 42.69   &26.66  & 35.54 &40.62 \\
& S+G+KP& 55.43  &  60.08  & 51.20 &59.94  &65.59   & 49.62  & 67.58   &37.24  & 59.76 &58.39 \\
& S+G+KP+T & 68.41  &  77.29  & 65.87 &73.34  &75.46   & 55.39  & 83.69   &46.83  & 75.32 &73.20 \\

 Mauwake & S & 32.40  &  41.63  & 33.41 &48.68  &46.22   & 35.48  & 39.83   &26.52  & 30.78 & 42.59\\
 & S+G& 38.61  &  45.57  & 38.05 &50.87  &51.32   & 34.36  & 46.20   &28.32  & 37.16 &44.15 \\
&  S+G+KP& 51.20  &  57.54  & 46.22 &59.54  &61.00   & 43.74  & 61.54   &36.34  & 51.65 &56.41 \\
& S+G+KP+T& 64.63  &  73.52  & 59.60 &71.24  &77.11   & 56.23  & 77.25   &43.59  & 73.05 &71.80  \\

 Kalamang  & S & 28.81  &  38.26  & 31.40 &39.33  &38.11   & 34.69  & 37.85   &27.44  & 32.47 &39.48 \\
 & S+G& 35.37  &  45.73  & 36.74 &41.92  &47.41   & 37.58  & 46.62   &28.35  & 37.04 &45.73 \\
 & S+G+KP& 62.20  &  65.19  & 52.74 &64.94  &67.99   & 54.71  & 68.77   &44.05  & 63.11&61.59  \\
 &  S+G+KP+T & 75.30  &  82.16  & 64.18 &77.90  &83.08   & 65.42  & 86.52   &54.42  & 80.55 &78.20  \\

Ulwa & S& 30.25  &  34.47  & 26.09 &41.22  &39.17   & 34.17  & 34.88   &27.61  & 30.09 &36.03 \\
& S+G & 34.25  &  42.86  & 30.09 &45.87  &48.68   & 33.50  & 39.23   &27.55  & 36.20 & 40.46 \\
 & S+G+KP& 53.92  &  61.78  & 44.41 &60.08  &64.34   & 46.28  & 66.39   &38.25  & 57.16 &55.11 \\
 & S+G+KP+T & 65.42  &  73.62  & 56.94 &72.18  &75.36   & 58.20  & 77.64   &45.43  & 73.80 &68.99 \\\hline

\multicolumn{12}{l}{\textbf{Indo-European Language Family}} \\
Palula & S & 28.73  &  31.06  & 27.42 &35.36  &35.19   & 32.66  & 32.71   &30.23  & 31.54 &35.96\\
 &  S+G & 38.47  &  40.28  & 37.22 &40.20  &43.25   & 32.87  & 43.73   &29.45  & 39.14 &36.68 \\
 & S+G+KP& 47.91  &  50.06  & 45.40 &47.97  &50.42   & 39.41  & 54.72   &34.95  & 52.15 &47.79 \\
 &  S+G+KP+T& 67.03  &  72.87  & 59.98 &68.04  &69.71   & 54.16  & 75.89   &44.92  & 73.78 &67.74  \\

Tuatschin & S & 29.29  &  35.13  & 29.02 &41.96  &33.69   & 31.60  & 40.53   &28.75  & 30.37 & 34.95 \\
 & S+G & 44.65  &  49.51  & 35.94 &44.65  &46.09   & 33.77  & 58.27   &29.83  & 45.64 &47.62 \\
& S+G+KP& 58.58  &  62.61  & 48.07 &60.74  &61.90   & 42.36  & 71.30   &34.86  & 63.43 &57.86  \\
& S+G+KP+T & 72.87  &  78.43  & 62.53 &73.41  &75.83   & 57.94  & 83.77   &47.26  & 79.34 &73.94\\\hline

\multicolumn{12}{l}{\textbf{Other Language Families}} \\
 Japhug& S & 27.65  &  31.01  & 31.28 &43.30  &45.81   & 32.34  & 36.49   &27.37  & 32.68 &36.87 \\
 & S+G & 38.83  &  51.40  & 40.50 &47.21  &53.91   & 30.33  & 48.86   &24.86  & 49.16 &48.60 \\
 & S+G+KP& 53.35  &  67.04  & 51.68 &63.13  &68.44   & 42.99  & 64.37   &37.71  & 61.90 &61.17 \\
 & S+G+KP+T & 66.20  &  81.28  & 67.88 &75.14  &80.73   & 54.27  & 83.10   &44.41  & 79.05 &73.18 \\

Yauyos Quechua & S & 32.98  &  38.06  & 31.06 &41.29  &39.11   & 31.64  & 35.19   &30.36  & 28.35 &43.74 \\
 & S+G & 37.27  &  42.91  & 31.41 &44.01  &40.86   & 33.43  & 43.25   &28.52  & 33.07 &39.81 \\
& S+G+KP& 53.42  &  57.62  & 47.51 &57.13  &57.66   & 50.66  & 62.79   &35.96  & 48.56 &53.11 \\
& S+G+KP+T & 71.92  &  80.40  & 64.36 &76.03  &78.65   & 61.34  & 84.82   &48.56  & 77.17 &75.24 \\

Mehweb & S & 30.59  &  25.88  & 20.00 &25.88  &30.59   & 25.00  & 34.12   &22.35  & 30.59 &24.71 \\
 & S+G & 31.76  &  32.94  & 25.88 &35.29  &44.71   & 21.43  & 40.24   &27.06  & 24.71 &24.71  \\
 & S+G+KP& 37.65  &  32.94  & 35.29 &34.12  &51.76   & 45.68  & 40.00   &20.00  & 30.59 &31.76 \\
 & S+G+KP+T& 60.00  &  63.53  & 44.71 &52.94  &70.59   & 55.00  & 65.48   &40.00  & 69.41 &51.76 \\

Ik & S & 28.57  &  28.57  & 33.33 &38.10  &23.81   & 40.00  & 33.33   &28.57  & 19.05 &38.10 \\
 &   S+G& 42.86  &  47.62  & 57.14 &38.10  &42.86   & 25.00  & 47.62   &38.10  & 42.86 &47.62 \\
 & S+G+KP& 57.14  &  57.14  & 57.14 &57.14  &66.67   & 30.00  & 80.95   &42.86  & 66.67 & 61.90 \\
 & S+G+KP+T & 90.48  &  95.24  & 85.71 &71.43  &90.48   & 76.47  & 95.24   &61.90  & 100.00 &90.48 \\
\bottomrule
\caption{Accuracy scores across languages and difficulties for all models.}
\label{tab:full-result-main-appendix}
\end{longtable}
\end{scriptsize}

\clearpage
\section{CoT and Non-CoT Prompting}
\label{sec:cot-full}
\begin{scriptsize}
\begin{longtable}{@{\extracolsep{\fill}}ll*{3}{S}}
 \\
\label{tab:cot-no-not-full} \\
\toprule
\textbf{Language} & \textbf{Difficulty} &
\multicolumn{1}{c}{\textbf{Qwen2.5-7B}} &
\multicolumn{1}{c}{\textbf{Gemma3-12B}} &
\multicolumn{1}{c}{\textbf{LLaMA3-8B}} \\
\midrule
\endfirsthead

\multicolumn{5}{c}{(\textit{Table \thetable{} continued from previous page})} \\
\toprule
\textbf{Language} & \textbf{Difficulty} &
\multicolumn{1}{c}{\textbf{Qwen2.5-7B}} &
\multicolumn{1}{c}{\textbf{Gemma3-12B}} &
\multicolumn{1}{c}{\textbf{LLaMA3-8B}} \\
\midrule
\endhead

\multicolumn{5}{r}{(\textit{Continued on next page})} \\
\endfoot
\endlastfoot

\multicolumn{5}{l}{\textbf{Atlantic-Congo Language Family}} \\
Pichi & S & 31.48 & 44.27 & 30.97 \\
 &  S+G & 46.03 & 49.51 & 37.15 \\
 & S+G+KP & 61.00 & 62.02 & 48.25 \\
 & S+G+KP+T & 73.86 & 75.86 & 60.26 \\

 & CoT-S & 36.47 & 40.02 & 34.80 \\
 & CoT-S+G & 43.68 & 45.71 & 45.52 \\
 & CoT-S+G+KP & 58.14 & 60.70 & 55.37 \\
 & CoT-S+G+KP+T & 71.10 & 77.20 & 70.28 \\\hline

Gyeli & S & 26.48 & 34.44 & 32.88 \\
 & S+G & 35.75 & 39.94 & 34.99 \\
 & S+G+KP & 55.57 & 60.35 & 47.54 \\
 & S+G+KP+T & 64.25 & 67.29 & 57.16 \\

 & CoT-S & 24.60 & 30.54 & 27.06 \\
 & CoT-S+G & 31.40 & 39.22 & 35.12 \\
 & CoT-S+G+KP & 54.27 & 56.67 & 53.48 \\
 & CoT-S+G+KP+T & 62.23 & 69.18 & 66.81 \\\hline

Moloko & S & 25.97 & 41.46 & 29.76 \\
 & S+G & 38.29 & 46.47 & 30.68 \\
 & S+G+KP & 51.94 & 59.68 & 46.42 \\
 & S+G+KP+T & 67.88 & 71.75 & 57.14 \\

 & CoT-S & 25.06 & 35.76 & 30.30 \\
 & CoT-S+G & 36.45 & 40.09 & 38.27 \\
 & CoT-S+G+KP & 50.57 & 53.76 & 49.20 \\
 & CoT-S+G+KP+T & 69.25 & 74.49 & 67.88 \\\hline

Fwe & S & 35.37 & 42.86 & 29.55 \\
 & S+G & 44.22 & 40.14 & 32.35 \\
 & S+G+KP & 59.18 & 57.14 & 53.44 \\
 & S+G+KP+T & 65.31 & 69.39 & 55.47 \\

 & CoT-S & 33.33 & 38.78 & 25.85 \\
 & CoT-S+G & 39.46 & 36.05 & 40.41 \\
 & CoT-S+G+KP & 53.74 & 53.06 & 48.63 \\
 & CoT-S+G+KP+T & 62.59 & 72.11 & 60.54 \\\hline\hline

\multicolumn{5}{l}{\textbf{Austronesian Language Family}} \\
Papuan Malay& S & 39.80 & 50.82 & 38.37 \\
  & S+G & 43.12 & 52.89 & 38.46 \\
  & S+G+KP & 57.89 & 63.78 & 48.28 \\
  & S+G+KP+T & 77.51 & 80.64 & 64.38 \\

  & CoT-S & 44.90 & 48.33 & 38.97 \\
  & CoT-S+G & 45.62 & 49.55 & 42.36 \\
  & CoT-S+G+KP & 56.51 & 62.79 & 51.77 \\
  & CoT-S+G+KP+T & 76.21 & 79.94 & 72.49 \\\hline

Rapa Nui & S & 38.68 & 43.89 & 28.38 \\
  & S+G & 51.84 & 46.58 & 35.19 \\
  & S+G+KP & 60.62 & 57.28 & 41.53 \\
  & S+G+KP+T & 73.96 & 70.39 & 52.59 \\

  & CoT-S & 34.82 & 34.11 & 35.25 \\
  & CoT-S+G & 44.12 & 42.36 & 44.05 \\
  & CoT-S+G+KP & 53.63 & 51.84 & 50.18 \\
  & CoT-S+G+KP+T & 65.75 & 69.20 & 64.51 \\\hline

Kagayanen & S & 30.36 & 39.82 & 30.26 \\
 & S+G & 43.64 & 41.82 & 32.45 \\
 & S+G+KP & 53.45 & 53.64 & 42.70 \\
 & S+G+KP+T & 66.73 & 69.82 & 55.70 \\

 & CoT-S & 32.91 & 38.00 & 33.64 \\
 & CoT-S+G & 39.45 & 42.91 & 40.55 \\
 & CoT-S+G+KP & 51.45 & 52.91 & 43.69 \\
 & CoT-S+G+KP+T & 66.18 & 69.64 & 64.84 \\\hline

Vamale & S & 31.34 & 43.28 & 39.06 \\
 & S+G & 34.33 & 50.57 & 38.46 \\
 & S+G+KP & 59.70 & 50.75 & 45.31 \\
 & S+G+KP+T & 68.66 & 71.64 & 63.64 \\

 & CoT-S & 19.40 & 31.34 & 35.82 \\
 & CoT-S+G & 29.85 & 37.31 & 38.81 \\
 & CoT-S+G+KP & 46.27 & 47.76 & 43.28 \\
 & CoT-S+G+KP+T & 61.19 & 68.66 & 70.15 \\\hline\hline

\multicolumn{5}{l}{\textbf{Trans-New Guinea Language Family}} \\
Komnzo  & S & 34.41 & 41.47 & 32.68 \\
 & S+G & 43.16 & 43.86 & 35.61 \\
 & S+G+KP & 55.43 & 59.94 & 49.62 \\
 & S+G+KP+T & 68.41 & 73.34 & 55.39 \\

 & CoT-S & 35.68 & 35.26 & 27.22 \\
 & CoT-S+G & 36.67 & 40.54 & 37.29 \\
 & CoT-S+G+KP & 53.17 & 57.97 & 48.30 \\
 & CoT-S+G+KP+T & 65.87 & 71.79 & 65.54 \\\hline

Mauwake  & S & 32.40 & 48.68 & 35.48 \\
 & S+G & 38.61 & 50.87 & 34.36 \\
 & S+G+KP & 51.20 & 59.54 & 43.74 \\
 & S+G+KP+T & 64.63 & 71.24 & 56.23 \\

 & CoT-S & 33.30 & 40.12 & 33.35 \\
 & CoT-S+G & 40.87 & 44.21 & 37.33 \\
 & CoT-S+G+KP & 50.14 & 57.58 & 47.56 \\
 & CoT-S+G+KP+T & 65.30 & 71.18 & 62.81 \\\hline

Kalamang & S & 28.81 & 39.33 & 34.69 \\
 & S+G & 35.37 & 41.92 & 37.58 \\
 & S+G+KP & 62.20 & 64.94 & 54.71 \\
 & S+G+KP+T & 75.30 & 77.90 & 65.42 \\

 & CoT-S & 34.76 & 33.84 & 32.01 \\
 & CoT-S+G & 39.63 & 37.96 & 37.00 \\
 & CoT-S+G+KP & 58.23 & 62.04 & 56.10 \\
 & CoT-S+G+KP+T & 74.09 & 78.96 & 73.02 \\\hline

Ulwa  & S & 30.25 & 41.22 & 34.17 \\
 & S+G & 34.25 & 45.87 & 33.50 \\
 & S+G+KP & 53.92 & 60.08 & 46.28 \\
 & S+G+KP+T & 65.42 & 72.18 & 58.20 \\

 & CoT-S & 32.41 & 35.66 & 31.37 \\
 & CoT-S+G & 37.33 & 40.36 & 36.26 \\
 & CoT-S+G+KP & 52.24 & 57.05 & 50.00 \\
 & CoT-S+G+KP+T & 65.59 & 71.91 & 64.25 \\\hline\hline
 
\multicolumn{5}{l}{\textbf{Indo-European Language Family}} \\
Palula  & S & 28.73 & 35.36 & 32.66 \\
 & S+G & 38.47 & 40.20 & 32.87 \\
 & S+G+KP & 47.91 & 47.97 & 39.41 \\
 & S+G+KP+T & 67.03 & 68.04 & 54.16 \\

 & CoT-S & 28.67 & 33.09 & 30.92 \\
 & CoT-S+G & 38.05 & 39.96 & 37.84 \\
 & CoT-S+G+KP & 45.10 & 45.94 & 46.24 \\
 & CoT-S+G+KP+T & 64.28 & 70.19 & 64.44 \\\hline

Tuatschin  & S & 29.29 & 41.96 & 31.60 \\
 & S+G & 44.65 & 44.65 & 33.77 \\
 & S+G+KP & 58.58 & 60.74 & 42.36 \\
 & S+G+KP+T & 72.87 & 73.41 & 57.94 \\

 & CoT-S & 31.09 & 38.85 & 32.43 \\
 & CoT-S+G & 45.01 & 46.45 & 42.59 \\
 & CoT-S+G+KP & 53.73 & 56.33 & 51.40 \\
 & CoT-S+G+KP+T & 67.30 & 73.05 & 68.10 \\\hline\hline

\multicolumn{5}{l}{\textbf{Other Language Families}} \\
Japhug  & S & 27.65 & 43.30 & 32.34 \\
 & S+G & 38.83 & 47.21 & 30.33 \\
 & S+G+KP & 53.35 & 63.13 & 42.99 \\
 & S+G+KP+T & 66.20 & 75.14 & 54.27 \\

 & CoT-S & 33.80 & 37.43 & 31.01 \\
 & CoT-S+G & 41.06 & 48.60 & 39.39 \\
 & CoT-S+G+KP & 55.03 & 65.36 & 51.13 \\
 & CoT-S+G+KP+T & 70.11 & 77.37 & 64.71 \\\hline\hline

Yauyos Quechua  & S & 32.98 & 41.29 & 31.64 \\
  & S+G & 37.27 & 44.01 & 33.43 \\
  & S+G+KP & 53.42 & 57.13 & 50.66 \\
  & S+G+KP+T & 71.92 & 76.03 & 61.34 \\

  & CoT-S & 34.12 & 35.61 & 33.83 \\
  & CoT-S+G & 39.37 & 39.63 & 34.33 \\
  & CoT-S+G+KP & 49.82 & 53.98 & 46.92 \\
  & CoT-S+G+KP+T & 69.73 & 77.87 & 66.55 \\\hline\hline

Mehweb & S & 30.59 & 25.88 & 25.00 \\
 & S+G & 31.76 & 35.29 & 21.43 \\
 & S+G+KP & 37.65 & 34.12 & 45.68 \\
 & S+G+KP+T & 60.00 & 52.94 & 55.00 \\

 & CoT-S & 36.90 & 25.88 & 29.41 \\
 & CoT-S+G & 30.12 & 34.12 & 32.94 \\
 & CoT-S+G+KP & 39.76 & 36.47 & 38.82 \\
 & CoT-S+G+KP+T & 62.35 & 55.29 & 45.88 \\\hline\hline

Ik & S & 28.57 & 38.10 & 40.00 \\
 & S+G & 42.86 & 38.10 & 25.00 \\
 & S+G+KP & 57.14 & 57.14 & 30.00 \\
 & S+G+KP+T & 90.48 & 71.43 & 76.47 \\

 & CoT-S & 23.81 & 19.05 & 33.33 \\
 & CoT-S+G & 42.86 & 42.86 & 47.62 \\
 & CoT-S+G+KP & 55.00 & 52.38 & 57.14 \\
 & CoT-S+G+KP+T & 90.48 & 76.19 & 90.48 \\

\bottomrule
\caption{Accuracy scores with and without CoT across languages and difficulties for select models.}
\end{longtable}
\end{scriptsize}

\clearpage
\section{Ablation Study}
\label{sec:allDifficultiesResults}

\begin{scriptsize}
\begin{longtable}{@{\extracolsep{\fill}}lll*{3}{l}}

\label{tab:allDifficultyScores} \\
\toprule
\textbf{Language} & \textbf{Difficulty} & 
\multicolumn{1}{l}{\textbf{Qwen2.5-7B}} &
\multicolumn{1}{l}{\textbf{Gemma3-12B}} &
\multicolumn{1}{l}{\textbf{DeepSeek-R1-7B}} &
\multicolumn{1}{l}{\textbf{LLaMA3-8B}} \\
\midrule
\endfirsthead
\multicolumn{6}{c}{(\textit{Table \thetable{} continued from previous page})} \\
\toprule
\textbf{Language} & \textbf{Prompt} & 
\multicolumn{1}{l}{\textbf{Qwen2.5-7B}} &
\multicolumn{1}{l}{\textbf{Gemma3-12B}} &
\multicolumn{1}{l}{\textbf{DeepSeek-R1-7B}} &
\multicolumn{1}{l}{\textbf{LLaMA3-8B}} \\
\midrule
\endhead
\multicolumn{6}{r}{(\textit{Continued on next page})} \\
\endfoot
\endlastfoot

% Atlantic-Congo family
\multicolumn{6}{l}{\textbf{Atlantic-Congo Language Family}} \\
Pichi & S & 31.48 & 44.27 & 30.97 & 30.29 \\
 & S+G & 46.03 & 49.51 & 37.15 & 35.10 \\
 & S+KP & 59.17 & 60.64 & 45.98 & 52.26 \\
 & S+T & 54.80 & 72.63 & 54.81 & 54.80 \\
 & S+G+KP & 61.00 & 62.02 & 48.25 & 41.78 \\
 & S+G+T & 71.54 & 72.49 & 57.30 & 50.67 \\
 & S+KP+T & 64.40 & 76.53 & 61.52 & 64.40 \\
 & S+G+KP+T & 73.86 & 75.86 & 60.26 & 53.51 \\
\cmidrule{1-6}

Gyeli & S & 26.48 & 34.44 & 32.88 & 30.68 \\
 & S+G & 35.75 & 39.94 & 34.99 & 20.09 \\
 & S+KP & 52.03 & 59.29 & 47.77 & 48.99 \\
 & S+T & 44.04 & 60.38 & 49.62 & 44.04 \\
 & S+G+KP & 55.57 & 60.35 & 47.54 & 39.36 \\
 & S+G+T & 58.61 & 61.22 & 48.94 & 38.64 \\
 & S+KP+T & 54.00 & 69.68 & 54.83 & 54.00 \\
 & S+G+KP+T & 64.25 & 67.29 & 57.16 & 47.03 \\
\cmidrule{1-6}

Moloko & S & 25.97 & 41.46 & 29.76 & 22.78 \\
 & S+G & 38.29 & 46.47 & 30.68 & 28.70 \\
 & S+KP & 50.00 & 57.54 & 41.18 & 42.73 \\
 & S+T & 48.06 & 68.38 & 51.54 & 48.06 \\
 & S+G+KP & 51.94 & 59.68 & 46.42 & 40.32 \\
 & S+G+T & 62.64 & 71.53 & 53.10 & 42.60 \\
 & S+KP+T & 61.78 & 76.12 & 57.11 & 61.78 \\
 & S+G+KP+T & 67.88 & 71.75 & 57.14 & 45.79 \\
\cmidrule{1-6}

Fwe & S & 35.37 & 42.86 & 29.55 & 29.93 \\
 & S+G & 44.22 & 40.14 & 32.35 & 29.25 \\
 & S+KP & 50.34 & 57.86 & 46.62 & 46.94 \\
 & S+T & 41.50 & 59.59 & 43.07 & 41.50 \\
 & S+G+KP & 59.18 & 57.14 & 53.44 & 36.73 \\
 & S+G+T & 55.78 & 57.82 & 40.56 & 42.86 \\
 & S+KP+T & 56.16 & 65.25 & 51.47 & 56.16 \\
 & S+G+KP+T & 65.31 & 69.39 & 55.47 & 48.98 \\\hline

% Austronesian family
\multicolumn{6}{l}{\textbf{Austronesian Language Family}} \\
Papuan Malay & S & 39.80 & 50.82 & 38.37 & 33.30 \\
 & S+G & 43.12 & 52.89 & 38.46 & 32.58 \\
 & S+KP & 58.55 & 60.64 & 48.03 & 49.43 \\
 & S+T & 53.72 & 74.51 & 60.01 & 53.72 \\
 & S+G+KP & 57.89 & 63.78 & 48.28 & 44.80 \\
 & S+G+T & 72.94 & 76.90 & 60.64 & 54.38 \\
 & S+KP+T & 65.09 & 79.68 & 64.19 & 65.09 \\
 & S+G+KP+T & 77.51 & 80.64 & 64.38 & 60.12 \\
\cmidrule{1-6}

Rapa Nui & S & 38.68 & 43.89 & 28.38 & 31.01 \\
 & S+G & 51.84 & 46.58 & 35.19 & 34.87 \\
 & S+KP & 49.21 & 52.46 & 39.64 & 43.75 \\
 & S+T & 46.20 & 65.96 & 47.28 & 46.20 \\
 & S+G+KP & 60.62 & 57.28 & 41.53 & 38.99 \\
 & S+G+T & 73.03 & 64.95 & 49.21 & 44.70 \\
 & S+KP+T & 55.98 & 68.60 & 51.65 & 55.98 \\
 & S+G+KP+T & 73.96 & 70.39 & 52.59 & 47.51 \\
\cmidrule{1-6}

Kagayanen & S & 30.36 & 39.82 & 30.26 & 25.64 \\
 & S+G & 43.64 & 41.82 & 32.45 & 32.18 \\
 & S+KP & 46.00 & 51.59 & 41.70 & 44.34 \\
 & S+T & 47.64 & 66.61 & 50.93 & 47.64 \\
 & S+G+KP & 53.45 & 53.64 & 42.70 & 39.09 \\
 & S+G+T & 66.00 & 66.00 & 52.06 & 47.82 \\
 & S+KP+T & 58.06 & 68.65 & 57.56 & 58.06 \\
 & S+G+KP+T & 66.73 & 69.82 & 55.70 & 52.73 \\
\cmidrule{1-6}

Vamale & S & 31.34 & 43.28 & 39.06 & 31.34 \\
 & S+G & 34.33 & 50.57 & 38.46 & 23.88 \\
 & S+KP & 49.25 & 59.09 & 43.94 & 46.27 \\
 & S+T & 39.39 & 60.61 & 61.90 & 39.39 \\
 & S+G+KP & 59.70 & 50.75 & 45.31 & 28.36 \\
 & S+G+T & 55.22 & 65.67 & 61.29 & 31.34 \\
 & S+KP+T & 61.19 & 57.63 & 60.61 & 54.34 \\
 & S+G+KP+T & 68.66 & 71.64 & 63.64 & 50.75 \\\hline

% Trans-New Guinea family
\multicolumn{6}{l}{\textbf{Trans-New Guinea Language Family}} \\
Komnzo & S & 34.41 & 41.47 & 32.68 & 26.09 \\
 & S+G & 43.16 & 43.86 & 35.61 & 26.66 \\
 & S+KP & 52.55 & 57.58 & 45.98 & 45.03 \\
 & S+T & 45.66 & 61.71 & 51.69 & 45.66 \\
 & S+G+KP & 55.43 & 59.94 & 49.62 & 37.24 \\
 & S+G+T & 62.06 & 67.14 & 51.41 & 37.52 \\
 & S+KP+T & 53.69 & 73.09 & 57.47 & 53.69 \\
 & S+G+KP+T & 68.41 & 73.34 & 55.39 & 46.83 \\
\cmidrule{1-6}

Mauwake & S & 32.40 & 48.68 & 35.48 & 26.52 \\
 & S+G & 38.61 & 50.87 & 34.36 & 28.32 \\
 & S+KP & 49.41 & 55.14 & 42.64 & 44.23 \\
 & S+T & 42.45 & 66.25 & 53.30 & 42.45 \\
 & S+G+KP & 51.20 & 59.54 & 43.74 & 36.34 \\
 & S+G+T & 60.88 & 68.72 & 54.34 & 40.91 \\
 & S+KP+T & 53.17 & 70.53 & 55.05 & 53.17 \\
 & S+G+KP+T & 64.63 & 71.24 & 56.23 & 43.59 \\
\cmidrule{1-6}

Kalamang & S & 28.81 & 39.33 & 34.69 & 27.44 \\
 & S+G & 35.37 & 41.92 & 37.58 & 28.35 \\
 & S+KP & 59.60 & 62.96 & 54.21 & 55.12 \\
 & S+T & 48.62 & 68.75 & 55.37 & 48.62 \\
 & S+G+KP & 62.20 & 64.94 & 54.71 & 44.05 \\
 & S+G+T & 64.02 & 67.07 & 58.82 & 42.23 \\
 & S+KP+T & 63.89 & 79.44 & 64.84 & 63.89 \\
 & S+G+KP+T & 75.30 & 77.90 & 65.42 & 54.42 \\
\cmidrule{1-6}

Ulwa & S & 30.25 & 41.22 & 34.17 & 27.61 \\
 & S+G & 34.25 & 45.87 & 33.50 & 27.55 \\
 & S+KP & 51.11 & 56.35 & 48.55 & 46.04 \\
 & S+T & 45.26 & 64.61 & 51.70 & 45.26 \\
 & S+G+KP & 53.92 & 60.08 & 46.28 & 38.25 \\
 & S+G+T & 58.18 & 66.18 & 48.50 & 37.06 \\
 & S+KP+T & 54.34 & 71.19 & 57.92 & 54.34 \\
 & S+G+KP+T & 65.42 & 72.18 & 58.20 & 45.43 \\\hline

% Indo-European family
\multicolumn{6}{l}{\textbf{Indo-European Language Family}} \\
Palula & S & 28.73 & 35.36 & 32.66 & 30.23 \\
 & S+G & 38.47 & 40.20 & 32.87 & 29.45 \\
 & S+KP & 43.84 & 46.43 & 40.09 & 37.67 \\
 & S+T & 49.01 & 66.01 & 52.27 & 49.01 \\
 & S+G+KP & 47.91 & 47.97 & 39.41 & 34.95 \\
 & S+G+T & 62.25 & 65.47 & 51.98 & 40.38 \\
 & S+KP+T & 55.66 & 67.16 & 55.85 & 55.66 \\
 & S+G+KP+T & 67.03 & 68.04 & 54.16 & 44.92 \\
\cmidrule{1-6}

Tuatschin & S & 29.29 & 41.96 & 31.60 & 28.75 \\
 & S+G & 44.65 & 44.65 & 33.77 & 29.83 \\
 & S+KP & 49.41 & 52.09 & 41.84 & 43.54 \\
 & S+T & 45.35 & 65.52 & 53.14& 45.35 \\
 & S+G+KP & 58.58 & 60.74 & 42.36 & 34.86 \\
 & S+G+T & 66.67 & 67.12 & 52.66 & 39.98 \\
 & S+KP+T & 57.78 & 69.88 & 56.89 & 57.78 \\
 & S+G+KP+T & 72.87 & 73.41 & 57.94 & 47.26 \\\hline

% Other language families
\multicolumn{6}{l}{\textbf{Other Language Families}} \\
Japhug & S & 27.65 & 43.30 & 32.34 & 27.37 \\
 & S+G & 38.83 & 47.21 & 30.33 & 24.86 \\
 & S+KP & 50.56 & 57.53 & 43.90 & 41.01 \\
 & S+T & 46.22 & 63.71 & 50.00& 46.22 \\
 & S+G+KP & 53.35 & 63.13 & 42.99 & 37.71 \\
 & S+G+T & 65.92 & 68.44 & 50.59 & 38.27 \\
 & S+KP+T & 58.43 & 72.73 & 51.80 & 58.43 \\
 & S+G+KP+T & 66.20 & 75.14 & 54.27 & 44.41 \\
\cmidrule{1-6}

Yauyos Quechua & S & 32.98 & 41.29 & 31.64 & 30.36 \\
 & S+G & 37.27 & 44.01 & 33.43 & 28.52 \\
 & S+KP & 51.49 & 53.76 & 50.23 & 41.01 \\
 & S+T & 46.45 & 72.00 & 56.27& 46.45 \\
 & S+G+KP & 53.42 & 57.13 & 50.66 & 35.96 \\
 & S+G+T & 66.23 & 69.82 & 55.28 & 43.39 \\
 & S+KP+T & 58.91 & 78.65 & 63.61 & 58.91 \\
 & S+G+KP+T & 71.92 & 76.03 & 61.34 & 48.56 \\
\cmidrule{1-6}

Mehweb & S & 30.59 & 25.88 & 25.00 & 22.35 \\
 & S+G & 31.76 & 35.29 & 21.43 & 27.06 \\
 & S+KP & 27.71 & 34.72 & 27.85 & 36.90 \\
 & S+T & 34.12 & 59.04 & 46.25 & 34.12 \\
 & S+G+KP & 37.65 & 34.12 & 45.68 & 20.00 \\
 & S+G+T & 56.47 & 55.29 & 50.62 & 43.53 \\
 & S+KP+T & 43.53 & 56.06 & 46.84 & 43.53 \\
 & S+G+KP+T & 60.00 & 52.94 & 55.00 & 40.00 \\
\cmidrule{1-6}

Ik & S & 28.57 & 38.10 & 40.00 & 28.57 \\
 & S+G & 42.86 & 38.10 & 25.00 & 38.10 \\
 & S+KP & 61.90 & 70.59 & 47.62& 61.90 \\
 & S+T & 66.67 & 80.95 & 61.90 & 66.67 \\
 & S+G+KP & 57.14 & 57.14 & 30.00 & 42.86 \\
 & S+G+T & 80.95 & 80.95 & 65.00 & 47.62 \\
 & S+KP+T & 94.74 & 82.35 & 80.00 & 94.74 \\
 & S+G+KP+T & 90.48 & 71.43 & 76.47 & 61.90 \\
 \bottomrule
 \caption{Accuracy scores across languages across all information permutations for selected models.} 
\end{longtable}

\end{scriptsize}

\clearpage
\section{Linguistic Subfields}
\label{sec:lingSubfieldResults}
\begin{scriptsize}
\begin{longtable}{@{\extracolsep{\fill}}
    >{\scriptsize}l  >{\scriptsize}l  
    *{2}{c}  *{3}{c}  *{2}{c}  *{2}{c}  c
}
\toprule
 &  & \multicolumn{2}{c}{\textbf{Qwen2.5}} 
      & \multicolumn{3}{c}{\textbf{Gemma 3}} 
      & \multicolumn{2}{c}{\textbf{DeepSeek-R1}} 
      & \multicolumn{2}{c}{\textbf{LLaMA3}} 
      & \multicolumn{1}{c}{\textbf{GPT-4}} \\
\cmidrule(lr){3-4} \cmidrule(lr){5-7} \cmidrule(lr){8-9} \cmidrule(lr){10-11} \cmidrule(lr){12-12}
\textbf{Subfield} & \textbf{Difficulty}
         & {7B} & {32B}
         & {4B} & {12B} & {27B}
         & {7B} & {32B}
         & {8B} & {70B}
         & {o4-mini} \\
\midrule
\endfirsthead

\multicolumn{12}{c}{\scriptsize\itshape Continued from previous page}\\
\toprule
 &  & \multicolumn{2}{c}{\textbf{Qwen2.5}} 
      & \multicolumn{3}{c}{\textbf{Gemma 3}} 
      & \multicolumn{2}{c}{\textbf{DeepSeek-R1}} 
      & \multicolumn{2}{c}{\textbf{LLaMA3}} 
      & \multicolumn{1}{c}{\textbf{GPT-4}} \\
\cmidrule(lr){3-4} \cmidrule(lr){5-7} \cmidrule(lr){8-9} \cmidrule(lr){10-11} \cmidrule(lr){12-12}
\textbf{Language} & \textbf{Difficulty}
         & {7B} & {32B}
         & {4B} & {12B} & {27B}
         & {7B} & {32B}
         & {8B} & {70B}
         & {o4-mini} \\
\midrule
\endhead

\midrule
\multicolumn{12}{r}{\scriptsize\itshape Continued on next page}\\
\endfoot

\endlastfoot

Morphology & S        & 33.90 & 41.20 & 34.47 & 46.81 & 44.04 & 36.65 & 40.86 & 29.79 & 31.77 & 41.49 \\
Morphology & S+G      & 40.28 & 44.36 & 39.22 & 48.65 & 46.95 & 37.47 & 47.06 & 30.21 & 37.87 & 42.69 \\
Morphology & S+G+KP   & 57.23 & 62.24 & 56.45 & 59.59 & 63.90 & 49.55 & 66.66 & 38.46 & 56.23 & 59.01 \\
Morphology & S+G+KP+T & 67.87 & 76.72 & 64.04 & 74.82 & 77.23 & 58.55 & 80.40 & 47.66 & 76.03 & 73.05 \\
Phonology  & S        & 33.80 & 35.21 & 42.25 & 30.99 & 28.17 & 27.61 & 44.49 & 36.62 & 39.44 & 36.62 \\
Phonology  & S+G      & 38.03 & 46.48 & 40.84 & 42.25 & 36.62 & 19.32 & 53.01 & 33.80 & 46.48 & 39.44 \\
Phonology  & S+G+KP   & 53.52 & 57.75 & 64.79 & 48.29 & 56.34 & 44.27 & 75.81 & 25.35 & 45.07 & 45.07 \\
Phonology  & S+G+KP+T & 71.83 & 83.10 & 59.16 & 66.20 & 74.65 & 55.84 & 85.53 & 57.75 & 77.47 & 73.24 \\
Pragmatics & S        & 28.06 & 38.13 & 30.21 & 37.41 & 39.57 & 28.12 & 38.13 & 28.78 & 32.37 & 33.10 \\
Pragmatics & S+G      & 41.73 & 44.60 & 35.97 & 35.97 & 44.61 & 26.99 & 39.47 & 31.65 & 43.89 & 41.01 \\
Pragmatics & S+G+KP   & 56.11 & 62.59 & 65.47 & 65.79 & 64.03 & 53.07 & 70.50 & 42.44 & 64.03 & 58.99 \\
Pragmatics & S+G+KP+T & 74.10 & 74.68 & 65.47 & 73.38 & 75.54 & 55.17 & 78.97 & 46.76 & 73.82 & 75.54 \\
Semantics  & S        & 33.71 & 38.78 & 34.44 & 45.40 & 39.81 & 33.24 & 38.22 & 28.44 & 32.99 & 40.95 \\
Semantics  & S+G      & 41.78 & 47.47 & 39.50 & 46.33 & 49.23 & 34.61 & 46.74 & 30.71 & 41.26 & 45.71 \\
Semantics  & S+G+KP   & 54.39 & 58.14 & 56.46 & 53.49 & 62.77 & 44.52 & 63.91 & 39.19 & 55.62 & 57.81 \\
Semantics  & S+G+KP+T & 66.29 & 73.52 & 58.64 & 69.60 & 74.77 & 56.48 & 78.05 & 47.98 & 74.10 & 71.04 \\
Syntax     & S        & 32.87 & 38.46 & 32.43 & 43.30 & 41.33 & 33.18 & 39.44 & 29.75 & 34.67 & 41.88 \\
Syntax     & S+G      & 41.76 & 47.01 & 38.80 & 47.05 & 48.25 & 35.18 & 48.36 & 30.84 & 42.85 & 46.32 \\
Syntax     & S+G+KP   & 56.13 & 61.09 & 60.28 & 56.39 & 61.62 & 46.08 & 65.36 & 39.60 & 60.21 & 57.19 \\
Syntax     & S+G+KP+T & 71.63 & 78.70 & 64.23 & 74.15 & 77.12 & 58.99 & 81.40 & 50.62 & 78.65 & 74.11 \\
\bottomrule
\caption{Accuracy scores across linguistic subfields and select difficulties.} 
\label{tab:subfieldScores} 
\end{longtable}
\end{scriptsize}

\clearpage
\section{Overview of Languages Processed}
\label{sec:languages}

\subsection{Language Families}
\label{subsec:languageFamilies}

\begin{table}[ht]
\centering
\begin{tabular}{
  >{\raggedright\arraybackslash}p{7cm}
  >{\raggedright\arraybackslash}p{4cm}
  >{\centering\arraybackslash}p{4cm}
}
\hline
\textbf{Reference Grammar Title} & \textbf{Language Family} & \textbf{Citation} \\
\hline
A grammar of Pichi & Atlantic-Congo & \citealp{pichi} \\
A grammar of Gyeli & Atlantic-Congo & \citealp{gyeli} \\
A grammar of Moloko & Atlantic-Congo & \citealp{moloko} \\
A grammar of Fwe & Atlantic-Congo & \citealp{fwe} \\

A grammar of Papuan Malay & Austronesian & \citealp{papuan_malay} \\
A grammar of Rapa Nui & Austronesian & \citealp{rapanui} \\
A grammar of Kagayanen & Austronesian & \citealp{kagayanen} \\
A grammar of Vamale & Austronesian & \citealp{vamale} \\

A grammar of Komnzo & Trans-New Guinea & \citealp{komnzo} \\
A grammar of Mauwake & Trans-New Guinea & \citealp{mauwake} \\
A grammar of Kalamang & Trans-New Guinea & \citealp{kalamang} \\
A grammar of Ulwa (Papua New Guinea) & Trans-New Guinea & \citealp{ulwa} \\

A grammar of Palula & Indo-European & \citealp{palula} \\
A grammar of Tuatschin & Indo-European & \citealp{tuatschin} \\

A grammar of Japhug & Sino-Tibetan & \citealp{japhug} \\
A grammar of Yauyos Quechua & Quechuan & 
\citealp{yauyos_quechua} \\
The Mehweb language & Northeast Caucasian & \citealp{mehweb} \\
The Ik language & Nilo-Saharan & \citealp{ik} \\
\hline
\end{tabular}
\caption{Reference grammars and their language families: We process reference grammars from the \textit{Studies in Diversity Linguistics} and \textit{Comprehensive Grammar Library} series published by \textit{Language Science Press}. Each language's corresponding language family and each chapter's linguistic subfield are determined by reading the relevant sections (shown in \ref{subsec:chapterCategorization}).}
\label{tab:grammars}
\end{table}

\subsection{Chapter Categorization}
\label{subsec:chapterCategorization}

\begin{longtable}{
  >{\raggedright\arraybackslash}p{3cm}
  >{\raggedright\arraybackslash}p{9cm}
  >{\centering\arraybackslash}p{3cm}
}

  \caption{Overview of extracted chapters by language and linguistic subfield, in the order they appear in their respective reference grammar.}
  \label{tab:chapterCategorization} \\
  \toprule
  \textbf{Language}   & \textbf{Chapter}  & \textbf{Subfield}    \\
  \midrule
  \endfirsthead
    \toprule
    \textbf{Language} & \textbf{Chapter}  & \textbf{Subfield}    \\
    \midrule
    \endhead
    \midrule
    \multicolumn{3}{r}{\textit{Continued on next page}}\\
    \endfoot
    
    \bottomrule
    \endlastfoot
    Pichi         & Introduction                                    & Other  \\
    Pichi         & Segmental phonology                             & Phonology  \\
    Pichi         & Suprasegmental phonology                        & Phonology  \\
    Pichi         & Morphology                                      & Syntax  \\
    Pichi         & The nominal system                              & Syntax  \\
    Pichi         & The verbal system                               & Syntax  \\
    Pichi         & The clause                                      & Syntax      \\
    Pichi         & Spatial and temporal relations                  & Syntax   \\
    Pichi         & Grammatical relations                           & Syntax      \\
    Pichi         & Clause linkage                                  & Syntax      \\
    Pichi         & Multiverb constructions                         & Syntax      \\
    Pichi         & Pragmatic elements and routines                 & Pragmatics   \\
    Pichi         & Pichi and Spanish in contact                    & Other       \\

    Gyeli         & Introduction                                    & Other  \\
    Gyeli         & Phonology                                       & Phonology  \\
    Gyeli         & Parts of speech                                 & Syntax  \\
    Gyeli         & Morphology                                      & Morphology  \\
    Gyeli         & The noun phrase                                 & Syntax      \\
    Gyeli         & The verbal complex                              & Syntax  \\
    Gyeli         & Simple clauses                                  & Syntax      \\
    Gyeli         & Complex clauses                                 & Syntax      \\

    Moloko        & Clause                                          & Syntax      \\
    Moloko        & The na marker and na constructions              & Syntax      \\
    Moloko        & Clause combining                                & Syntax      \\
    Moloko        & Grammatical classes                             & Syntax  \\
    Moloko        & Noun morphology                                 & Morphology  \\
    Moloko        & Noun phrase                                     & Syntax      \\
    Moloko        & The verb complex                                & Syntax      \\
    Moloko        & Verb phrase                                     & Syntax      \\
    Moloko        & Verb types and transitivity                     & Syntax      \\

    Fwe           & Mood                                            & Semantics  \\
    Fwe           & Negation                                        & Semantics   \\
    Fwe           & Syntax and information structure                & Syntax      \\
    Fwe           & Nominal morphology                              & Morphology  \\
    Fwe           & Minor word categories                           & Syntax  \\
    Fwe           & Verbal derivation                               & Syntax  \\
    Fwe           & Tense                                           & Semantics  \\

    Papuan Malay   & Introduction                                    & Other  \\
    Papuan Malay   & Phonology                                       & Phonology  \\
    Papuan Malay   & Word-formation                                  & Morphology  \\
    Papuan Malay   & Reduplication                                   & Morphology  \\
    Papuan Malay   & Word classes                                    & Syntax  \\
    Papuan Malay   & Personal pronouns                               & Syntax  \\
    Papuan Malay   & Demonstratives and locatives                    & Syntax  \\
    Papuan Malay   & Noun phrases                                    & Syntax      \\
    Papuan Malay   & Adnominal possessive relations                  & Syntax      \\
    Papuan Malay   & Prepositions and the prepositional phrase       & Syntax      \\
    Papuan Malay   & Verbal clauses                                  & Syntax      \\
    Papuan Malay   & Nonverbal clauses                               & Syntax      \\
    Papuan Malay   & Negative, interrogative, and directive clauses  & Syntax      \\
    Papuan Malay   & Conjunctions and constituent combining          & Syntax      \\

    Rapa Nui       & Introduction                                    & Other  \\
    Rapa Nui       & Nouns and verbs                                 & Syntax  \\
    Rapa Nui       & Closed word classes                             & Syntax  \\
    Rapa Nui       & Noun phrase                                     & Syntax      \\
    Rapa Nui       & Possession                                      & Syntax      \\
    Rapa Nui       & Verb phrase                                     & Syntax      \\
    Rapa Nui       & Verbal clause                                   & Syntax      \\
    Rapa Nui       & Nonverbal clauses                               & Syntax      \\
    Rapa Nui       & Mood                                            & Semantics  \\
    Rapa Nui       & Combining clauses                               & Syntax      \\

    Kagayanen     & Voice                                           & Syntax  \\
    Kagayanen     & Pragmatically marked structures                 & Pragmatics   \\
    Kagayanen     & Clause combining                                & Syntax      \\
    Kagayanen     & Referring expressions                           & Semantics   \\
    Kagayanen     & Modification                                    & Semantics      \\
    Kagayanen     & Non-verbal clauses                              & Syntax      \\
    Kagayanen     & Verb structure and inflection                   & Syntax  \\
    Kagayanen     & Stem-forming morphological processes            & Morphology  \\
    Kagayanen     & Morphosyntactically defined verb classes        & Syntax  \\
    Kagayanen     & Semantically motivated verb classes             & Semantics   \\

    Vamale        & Noun phrases                                    & Syntax      \\
    Vamale        & Nouns                                           & Syntax  \\
    Vamale        & Verb phrases                                    & Syntax      \\
    Vamale        & Verbs                                           & Syntax  \\
    Vamale        & Voice                                           & Syntax  \\
    Vamale        & Word classes                                    & Syntax  \\

    Komnzo        & Word classes                                    & Syntax  \\
    Komnzo        & Nominal morphology                              & Morphology  \\
    Komnzo        & Verb morphology                                 & Morphology  \\
    Komnzo        & Tense, aspect and mood                          & Semantics  \\
    Komnzo        & Syntax of the noun phrase                       & Syntax      \\
    Komnzo        & Clausal syntax                                  & Syntax      \\
    Komnzo        & Complex syntax                                  & Syntax      \\
    Komnzo        & Aspects of the lexicon                          & Semantics   \\

    Mauwake       & Introduction                                    & Other  \\
    Mauwake       & Morphology                                      & Morphology  \\
    Mauwake       & Phrase level syntax                             & Syntax      \\
    Mauwake       & Clause                                          & Syntax      \\
    Mauwake       & Functional domains                              & Semantics   \\
    Mauwake       & Sentence types                                  & Syntax      \\
    Mauwake       & Clause combinations                             & Syntax      \\
    Mauwake       & Theme, topic, and focus                         & Semantics   \\

    Kalamang      & Morphological units and processes               & Morphology  \\
    Kalamang      & Word classes                                    & Syntax  \\
    Kalamang      & Nouns, noun phrases and postpositional phrases  & Syntax  \\
    Kalamang      & Pronouns and person reference and address       & Syntax  \\
    Kalamang      & Quantifiers                                     & Semantics  \\
    Kalamang      & Possessive and associative constructions        & Semantics      \\
    Kalamang      & Demonstratives                                  & Semantics  \\
    Kalamang      & Verbs                                           & Syntax  \\
    Kalamang      & The clause                                      & Syntax      \\
    Kalamang      & Complex predicates                              & Syntax      \\
    Kalamang      & Clausal modification                            & Syntax      \\
    Kalamang      & Multiclausal constructions                      & Syntax      \\
    Kalamang      & Information structure                           & Syntax   \\
    Kalamang      & Other topics                                    & Other       \\

    Ulwa          & Adjectives                                      & Syntax  \\
    Ulwa          & Clause-level syntax                             & Syntax      \\
    Ulwa          & Complex sentences                               & Syntax      \\
    Ulwa          & Determiners                                     & Syntax  \\
    Ulwa          & The structural consequences of language loss    & Syntax       \\
    Ulwa          & The Maruat-Dimiri-Yaul dialect of Ulwa          & Other       \\
    Ulwa          & Nouns                                           & Syntax  \\
    Ulwa          & Other word classes                              & Syntax  \\
    Ulwa          & A grammatical overview of Ulwa                  & Syntax       \\
    Ulwa          & Phrase-level syntax                             & Syntax      \\
    Ulwa          & Predicates                                      & Syntax      \\
    Ulwa          & Pronouns                                        & Syntax  \\
    Ulwa          & Topics in semantics                             & Semantics   \\
    Ulwa          & Additional topics in syntax                     & Syntax      \\
    Ulwa          & Verbs                                           & Syntax  \\

    Palula        & Typological overview                            & Other       \\
    Palula        & Nouns                                           & Syntax  \\
    Palula        & Pronouns                                        & Syntax  \\
    Palula        & Adjectives and quantifiers                      & Syntax  \\
    Palula        & Adverbs and postpositions                       & Syntax  \\
    Palula        & Verbs                                           & Syntax  \\
    Palula        & Verbal categories                               & Syntax  \\
    Palula        & Noun phrases and non-verbal agreement           & Syntax      \\
    Palula        & Grammatical relations                           & Syntax      \\
    Palula        & Simple clauses and argument structure           & Syntax      \\
    Palula        & Complex constructions                           & Syntax      \\
    Palula        & Sentence modification                           & Syntax      \\

    Tuatschin     & Phonology                                       & Phonology   \\
    Tuatschin     & Noun phrase                                     & Syntax      \\
    Tuatschin     & Verb phrase                                     & Syntax      \\
    Tuatschin     & Simple sentences                                & Syntax      \\
    Tuatschin     & Complex sentences                               & Syntax      \\
    Tuatschin     & Morphological processes                         & Morphology  \\
    
    Japhug        & A grammatical sketch                            & Syntax       \\
    Japhug        & Phonology                                       & Phonology    \\
    Japhug        & Nominal morphology                              & Morphology  \\
    Japhug        & Pronouns                                        & Syntax  \\
    Japhug        & Postpositions and relator nouns                 & Syntax  \\
    Japhug        & The noun phrase                                 & Syntax      \\
    Japhug        & Expressive words and sentence final particles   & Syntax  \\
    Japhug        & Non-concatenative verbal morphology             & Morphology  \\
    Japhug        & Tense, aspect, modality and evidentiality       & Semantics  \\
    Japhug        & Simple clauses                                  & Syntax      \\
    Japhug        & Relative clauses                                & Syntax      \\
    Japhug        & Complement clauses                              & Syntax      \\
    Japhug        & Other types of multiclausal constructions       & Syntax      \\
    Japhug        & Degree and comparison                           & Semantics  \\
    
    Yauyos Quechua & Substantives                                    & Syntax  \\
    Yauyos Quechua & Verbs                                           & Syntax  \\
    Yauyos Quechua & Particles                                       & Syntax  \\
    Yauyos Quechua & Enclitics                                       & Syntax  \\
    Yauyos Quechua & Syntax                                          & Syntax      \\
    Yauyos Quechua & Further analysis of evidential modifiers        & Syntax  \\

    Mehweb        & Phonology                                       & Phonology  \\
    Mehweb        & Mood of Mehweb                                  & Semantics  \\
    Mehweb        & Causatives                                      & Syntax      \\
    Mehweb        & Assertive copula in Mehweb                      & Other      \\
    
    Ik            & Adverbs                                         & Syntax  \\
    Ik            & Case                                            & Syntax  \\
    Ik            & Demonstratives                                  & Semantics  \\
    Ik            & Morphology                                      & Morphology  \\
    Ik            & Verbs                                           & Syntax  \\

\end{longtable}

\clearpage
\section{A Concrete Prompt Example}
% See Figure \ref{fig:examplefullPromptTemplate}
\label{sec:concrete}
\begin{figure}[ht]
        \begin{tcolorbox}[
        enhanced,
        colback=white,
        colframe=black,
        arc=3mm,
        boxrule=0.5pt,
        title=Full Prompt Example
        ]
        {You are a linguist specializing in Palula. You are given a sentence along with its morpheme breakdown}, 
        {gloss}, and 
        {translation}. 
        {Words are separated by spaces, and morphemes are separated by hyphens. However, a word } 
        {and its gloss} 
        {are missing and represented by an underscore. Based on your understanding, please choose the most appropriate option.}
        
        \vspace{1ex}
        \textbf{{Sentence (with missing item)}}: pan\v{j} phuṭ-í \underline{\hspace{1cm}} phuṭ-í kir dít-u síinta.
        
        \vspace{1ex}
        \textbf{Gloss (with missing item)}: five  foot-PL \underline{\hspace{1cm}} foot-PL snow fall.PFV=MSG CONDH
        
        \vspace{1ex}
        \textbf{The English translation of this sentence is}: ‘When five or XXX feet snow had fallen...’
        
        \vspace{1ex}
        \textbf{{Here is a relevant knowledge point for this example, with the related morphemes}
        and glosses masked}: Another strategy for quantification, is by means of a partitive noun phrase. It specifies the quantity of the head noun, often itself preceded by or modified by a~cardinal numeral. Typically, but not exclusively, the nouns used in such partitive phrases denote containers or measuring terms of various kinds. In many ways it would make sense to describe higher numerals (such as 20, 100, 1000) as heads of partitive phrases, modified by the cardinal numerals 1--19 to express the numbers 21--39, etc.
        \vspace{2ex}
        
        \textbf{Options:}\\
        A: word: dubhiš=ee=ṣoṛíiš	 gloss: two.twenty=and=sixteen\\
        B: word: xálak-a	 gloss: people-PL\\
        C: word: ṣo	 gloss: six\\
        D: word: xálaka		 gloss: people
        
        \vspace{1ex}
        
        Please only return the letter (A–D). Do not output anything else.\\

        \textbf{DeepSeek-R1-7B result: C} \\
        
        \textbf{Correct Answer: C}
        \end{tcolorbox}
    \caption{A full prompt example of Palula and its prediction under the S+G+KP+T setting.}
    \label{fig:examplefullPromptTemplate}
\end{figure}

\end{document}